\documentclass{article}

\usepackage[final,nonatbib]{neurips_2021}

\usepackage[utf8]{inputenc} 
\usepackage[T1]{fontenc}    
\usepackage{hyperref}       
\usepackage{url}            
\usepackage{booktabs}       
\usepackage{amsfonts}       
\usepackage{nicefrac}       
\usepackage{microtype}      
\usepackage{xcolor}     

\usepackage{graphicx}
\usepackage{subfigure}
\usepackage{wrapfig}
\usepackage[normalem]{ulem}
\usepackage{enumitem}

\usepackage{amsmath,amssymb,amsfonts}
\usepackage{multirow}
\usepackage{pifont}

\usepackage[normalem]{ulem}

\usepackage[ruled,vlined]{algorithm2e}

\SetCommentSty{mycommfont}

\SetKwInput{KwInput}{Input}                
\SetKwInput{KwOutput}{Output}   
\SetKwInput{KwInit}{Initialization}  
\SetKwComment{Comment}{$\triangleright$\ }{}

\renewcommand\footnotemark{}

\title{MEST: Accurate and Fast Memory-Economic Sparse Training Framework on the Edge}

\author{%
  Geng Yuan\text{$^{1,\dagger}$}\thanks{$\dag$ These authors contributed equally.},
  Xiaolong Ma\text{$^{1,\dagger}$},
  Wei Niu\text{$^{2}$}, 
  Zhengang Li\text{$^{1}$}, 
  Zhenglun Kong\text{$^{1}$}, 
  Ning Liu\text{$^{3}$}, \vspace{0.4em} \\ 
  \textbf{Yifan Gong\text{$^{1}$},} 
  \textbf{Zheng Zhan\text{$^{1}$}, }
  \textbf{Chaoyang He\text{$^{4}$}, }
  \textbf{Qing Jin\text{$^{1}$}, }
  \textbf{Siyue Wang\text{$^{1}$}, } \vspace{0.4em} \\ 
  \textbf{Minghai Qin, }
  \textbf{Bin Ren\text{$^{2}$}, }
  \textbf{Yanzhi Wang\text{$^{1}$}, }
  \textbf{Sijia Liu\text{$^{5}$}, }
  \textbf{Xue Lin\text{$^{1}$} } \vspace{0.4em} \\ 
  \text{$^{1}$} Northeastern University, 
  \text{$^{2}$} College of William and Mary, \\
  \text{$^{3}$} Midea Group,
  \text{$^{4}$} University of Southern California, 
  \text{$^{5}$} Michigan State University \\
  \small{\texttt{\{yuan.geng,xue.lin\}@northeastern.edu}}
}

\begin{document}

\maketitle

\begin{abstract}
Recently, a new trend of exploring sparsity for accelerating neural network training has emerged, embracing the paradigm of training on the edge.
This paper proposes a novel Memory-Economic Sparse Training (MEST) framework targeting for accurate and fast execution on edge devices.
The proposed MEST framework consists of enhancements by Elastic Mutation (EM) and Soft Memory Bound (\&S) that ensure superior accuracy at high sparsity ratios.
Different from the existing works for sparse training, this current work
reveals the importance of \emph{sparsity schemes} on the performance of sparse training in terms of accuracy as well as training speed on real edge devices.
On top of that, the paper proposes to employ data efficiency for further acceleration of sparse training. 
Our results suggest that unforgettable examples can be identified  \emph{in-situ} even during the dynamic exploration of sparsity masks in the sparse training process, and therefore can be removed for further training speedup on edge devices.
Comparing with state-of-the-art (SOTA) works on accuracy, our MEST increases Top-1 accuracy significantly on ImageNet when using the same unstructured sparsity scheme.
Systematical evaluation on accuracy, training speed, and memory footprint are conducted, where the proposed MEST framework consistently  outperforms representative SOTA works. Our codes are publicly available at: \url{https://github.com/boone891214/MEST}.

\end{abstract}

\newcommand{\GY}[1]{\textcolor{blue}{Geng: #1}}

\newcommand{\todo}[1]{\textcolor{red}{\sf\bfseries Todo: #1}}

\newcommand{\bred}[1]{\textcolor{red}{\sf\bfseries #1}}
\newcommand{\red}[1]{\textcolor{red}{#1}}

\newcommand{\blue}[1]{\textcolor{blue}{#1}}

\newcommand{\xl}[1]{\textcolor{orange}{#1}}

\newcommand{\cross}[1]{\textcolor{red}{\sout{#1}}}

\section{Introduction}\label{sec:intro}
To promote the broader applications of deep learning on the edge, a surge of research efforts have been devoted to removing the over-parameterization in neural networks for accelerated inference.
Specifically, existing works have explored various strategies such as heuristics-based pruning \cite{guo2016dynamic,yu2018nisp}, regularization-based pruning \cite{wen2016learning,liu2018rethinking}, and recently prevailing network architecture search \cite{zoph2016neural,bender2018understanding,pham2018efficient,real2019regularized}.

Recently, a new trend of exploring sparsity for training acceleration of neural networks has emerged to embrace the promising training-on-the-edge paradigm.
The first works in this direction use the pruning-at-initialization approach such as SNIP \cite{lee2018snip} and GraSP \cite{wang2019picking} that first obtains a fixed sparse model structure and then follows with a traditional training process.
However, the whole process is still computation- and memory-intensive, and therefore not compatible with the  end-to-end edge training paradigm. 
{Such a sparse training methodology with the pre-fixed structure also faces the problem of compromised accuracy.}

\begin{wrapfigure}{R}{0.5\textwidth}
    \centering
    \includegraphics[width=0.5 \textwidth]{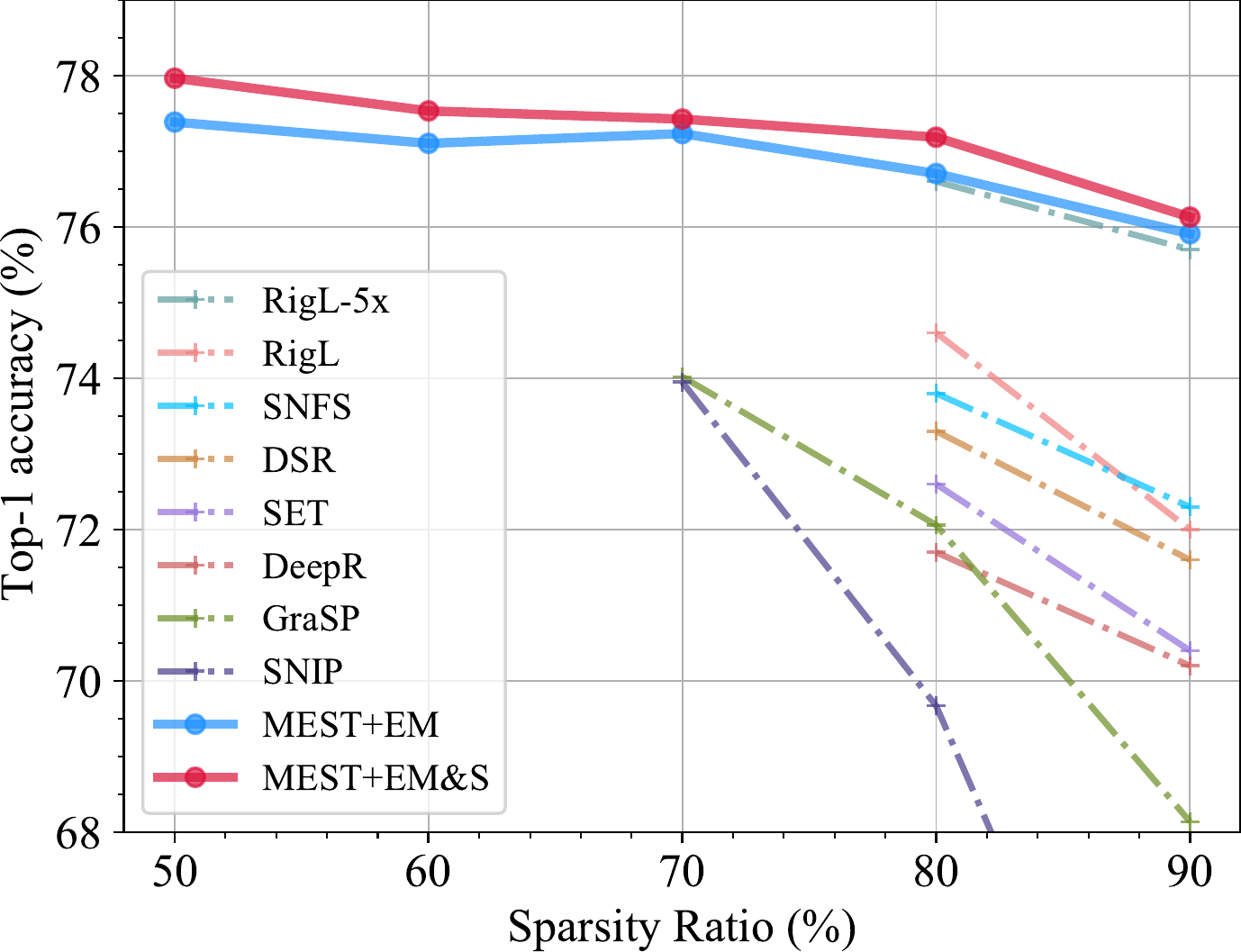}
    \caption{Accuracy vs sparsity ratio on ImageNet using ResNet-50 dense model. Our proposed MEST framework: MEST+EM (Elastic Mutation) and MEST+EM\&S (with Soft Memory Bound) are compared with the SOTA sparse training algorithms i.e., GraSP~\cite{wang2019picking}, SNIP~\cite{lee2018snip}, RigL~\cite{evci2020rigging}, SNFS~\cite{dettmers2019sparse}, DSR~\cite{mostafa2019parameter}, SET~\cite{mocanu2018scalable}, and DeepR~\cite{bellec2018deep}.
    }
    \label{fig:imagenet_acc_vs_prune}
\vspace{-0.5em}
\end{wrapfigure}

Furthermore, sparse training with dynamic sparsity mask such as SET \cite{mocanu2018scalable}, DeepR \cite{bellec2018deep}, DSR \cite{mostafa2019parameter},
RigL \cite{evci2020rigging}, 
and SNFS \cite{dettmers2019sparse} have been proposed, showing 
great potential towards end-to-end edge training. 
Specifically, they start with a sparse model structure picked intuitively from the initialized (but not trained) dense model, and then heuristically explore various sparse topologies at the given sparsity ratio, together with the sparse model training.
The underlying principle of sparse training is that the total epoch number is the same as  dense training, but the speed of each training iteration (batch) is significantly improved, thanks to the reduced computation amount due to sparsity. 

This paper proposes a novel Memory-Economic Sparse Training (MEST) framework targeting for accurate and fast execution on edge devices.
Specifically, we boost the accuracy with the MEST+EM (Elastic Mutation) to effectively explore various sparse topologies.
And we propose another enhancement through applying a soft memory bound, namely, MEST+EM\&S, which relaxes on the memory footprint during sparse training with the target sparsity ratio met by the end of training.
In Figure \ref{fig:imagenet_acc_vs_prune}, the accuracy of the proposed MEST is compared against state-of-the-art (SOTA) sparse training algorithms under different sparsity ratios, using the same unstructured sparsity scheme.

Furthermore, as with inference acceleration, we find that sparse training closely relates to the adopted sparsity scheme such as unstructured \cite{han2015deep}, structured \cite{molchanov2017variational,dai2018compressing}, or fine-grained structured \cite{yang2018efficient} scheme, which can result in varying accuracy, training speed, and memory footprint performance for sparse training.
With our effective MEST framework, this paper systematically investigates the sparse training problem with respect to the sparsity schemes. 
Specifically, besides the directly observable model accuracy, we conduct thorough analysis on the memory footprint by different sparsity schemes during sparse training, and in addition, {we  measure the training speed performance under various schemes on  the mobile edge device.}

On top of that, the paper proposes to employ data efficiency for further acceleration of sparse training. 
The prior works~\cite{bengio2007scaling,kumar2010self,fan2017learning,toneva2018empirical} show that the amount of information provided by each training example is different, and the hardness of having an example correctly learned is also different.
Some training examples can be learned correctly early in the training stage and will never be ``forgotten'' (i.e., misclassified) again.
And removing those easy and less informative examples from the training dataset will not cause accuracy degradation on the final model.
However, the research of connecting training data efficiency to a sparse training scene is still missing, due to the dynamic sparsity mask.
In this work, we explore the impact of model sparsity, sparsity schemes, and sparse training algorithm on the amount of removable training examples.
And we also propose a data-efficient two-phase sparse training approach to effectively accelerate sparse training by removing less informative examples during the training process without harming the final accuracy.
The contributions of this work are summarized as follows:

\begin{itemize}[leftmargin=*]
    \item A novel Memory-Economic Sparse Training (MEST) framework with enhancements by Elastic Mutation (EM) and the soft memory bound targeting for accurate and fast execution on the edge. 
    \item A systematic investigation of the impact of sparsity scheme on the accuracy, memory footprint, as well as training speed with real edge devices, providing guidance for future edge training paradigm.
    \item Exploring the training data efficiency in the sparse training scenario for further training acceleration, by proposing a two-phase sparse training approach for \emph{in-situ} identification and removal of less informative examples during the training without hurting the final accuracy.
    \item  On CIFAR-100 with ResNet-32, comparing with representative SOTA sparse training works, i.e., SNIP, GraSP, SET, and DSR, our MEST increases accuracy by 1.91\%, 1.54\%, 1.14\%, and 1.17\%; 
achieves 1.76$\times$, 1.65$\times$, 1.87$\times$, and 1.98$\times$ training acceleration rate; 
and reduces the memory footprint by 8.4$\times$, 8.4$\times$, 1.2$\times$, and 1.2$\times$, respectively.

\end{itemize}

\section{Background and Related Work}\label{sec:background}
This section introduces representative neural network sparsity schemes and their impacts on memory footprint in sparse training, as well as the neural network sparse training strategies.

\subsection{Sparsity Scheme}
Neural network pruning has been well investigated for inference acceleration. 
The majority of works in this direction apply a pretraining-pruning-retraining flow, which is not compatible with the training-on-the-edge paradigm.
According to the adopted sparsity scheme, those works can be categorized as \emph{unstructured}~\cite{han2015deep,guo2016dynamic}, \emph{structured}~\cite{luo2017thinet,yu2018nisp,dong2019network,hu2016network,molchanov2017variational,wen2016learning,he2017channel,he2018soft,li2019compressing,he2019filter,neklyudov2017structured,dai2018compressing,zhang2021structadmm,ma2021non}, and \emph{fine-grained structured}~\cite{yang2018efficient,ma2020pconv,niu2020patdnn,ma2020image,dong2020rtmobile,rumi2020accelerating,zhang2021click, niu2021grim,guan2021cocopie} including the pattern-based and block-based ones. Detailed discussion about these sparsity schemes is provided in Appendix~\ref{sec:appen_scheme}.

Although these sparsity schemes are mainly proposed for accelerating inference, we find that they also play an important role in sparse training in terms of accuracy, memory footprint, and training speed.
For memory footprint, we focus on the two major components that vary with the sparsity scheme: \emph{model representation} together with \emph{gradients} produced during training. The detailed analysis is provided in Appendix~\ref{sec:appen_memory}.

\vspace{-0.5em}
\subsection{Sparse Training}
\vspace{-0.5em}
Majority of the sparse training works can be categorized into two groups: fixed-mask sparse training and dynamic-mask sparse training.
Additionally, there exist works \cite{lym2019prunetrain,you2020drawing,rajpal2020balancing} that prune dense networks in the early training stage, but they are out of scope for sparse training on the edge.

\vspace{-0.5em}
\subsubsection{Sparse Training with Fixed Sparsity Mask}
\vspace{-0.5em}
The fixed-mask approach \cite{lee2018snip,wang2019picking,tanaka2020pruning,wimmer2020freezenet,van2020single} has been proposed to decouple pruning and training such that after pruning, the sparse model training can be executed on edge devices.
SNIP~\cite{lee2018snip} preserves the loss after pruning based on connection sensitivity. 
GraSP~\cite{wang2019picking} prunes connections in a way that accounts for their role in the network’s gradient flow.
SynFlow \cite{tanaka2020pruning} 
proposes iterative synaptic flow pruning, which avoids layer collapse and preserves the total flow of synaptic strengths throughout the network. 
Since the proposed pruning algorithm does not incorporate back propagation, it achieves global pruning at initialization without data.
Based on the unstructured SNIP objective, 3SP~\cite{van2020single} further introduces a computation-aware weighting of the pruning score. 
This actively biases pruning by removing more computation-intensive channels which either have small effect on the loss or have  significant computation cost.
However, the pruning-at-initialization process is still computation and memory-intensive and therefore not compatible with the end-to-end sparse training on the edge.
And these works (except 3SP) employ the unstructured sparsity scheme.

\vspace{-0.5em}
\subsubsection{Sparse Training with Dynamic Sparsity Mask}
\vspace{-0.5em}
To reduce the computation as well as memory footprint during the whole training phase, sparse training is exploited in many works~\cite{bellec2018deep,mocanu2018scalable,mostafa2019parameter,dettmers2019sparse,evci2020rigging}, which can adjust the sparsity topology during training as well as maintain a low memory footprint.
Sparse Evolutionary Training (SET)~\cite{mocanu2018scalable} uses magnitude-based pruning and random growth at the end of each training epoch.
DeepR~\cite{bellec2018deep} combines dynamic sparse parameterization with stochastic parameter updates for training.
This method is primarily demonstrated with sparsification of fully-connected layers and applied to relatively small and shallow networks.
DSR~\cite{mostafa2019parameter} develops a dynamic reparameterization method to achieve high parameter efficiency in training sparse deep residual networks.
SNFS~\cite{dettmers2019sparse} develops sparse momentum, an algorithm which uses exponentially smoothed gradients (momentum) to identify layers and weights which reduce the error efficiently.
RigL~\cite{evci2020rigging} proposes to iteratively update sparse model topology during training by calculating dense gradients only at the update step.
Note that, though SNFS and RigL are sparse training, they actually involve computation of all the gradients corresponding to both pruned and non-zero weights, and therefore their memory footprint is equivalent to that of dense training.

\vspace{-0.5em}
\section{Sparse Training on the Edge}
\vspace{-0.5em}

\subsection{Sparsity Scheme in Sparse Training on the Edge}
\vspace{-0.5em}

It is common to see that sparse training works~\cite{lee2018snip,wang2019picking,bellec2018deep,mocanu2018scalable,mostafa2019parameter,dettmers2019sparse,evci2020rigging,tanaka2020pruning} represent the training speed performance using the training FLOP count.
Actually, such FLOPS cannot account for the actual execution overheads caused by the sparse data operations.
For example, the unstructured sparsity exhibits an irregular memory access pattern, leading to significant execution overhead.
Moreover, the dense model can take advantage of Winograd~\cite{lavin2016fast} to significantly accelerate the computation speed, which may apply in sparse computation.
Therefore, this paper directly measures the training speed performance on a mobile device.
We investigate the training acceleration performance by different sparsity schemes including unstructured, structured, and two state-of-the-art fine-grained structured (i.e., block and pattern), through a prototype implementation on a mobile edge device.
To do so, we conduct compiler-level  optimizations, leveraging a computation-graph based compilation approach, including optimizations on computation graph itself, tensor optimizations, etc.
More details are provided in Appendix~\ref{sec:appen_compiler}.

\begin{wrapfigure}{R}{0.5\textwidth}
    \vspace{-1em}
    \centering
    \includegraphics[width=0.5 \textwidth]{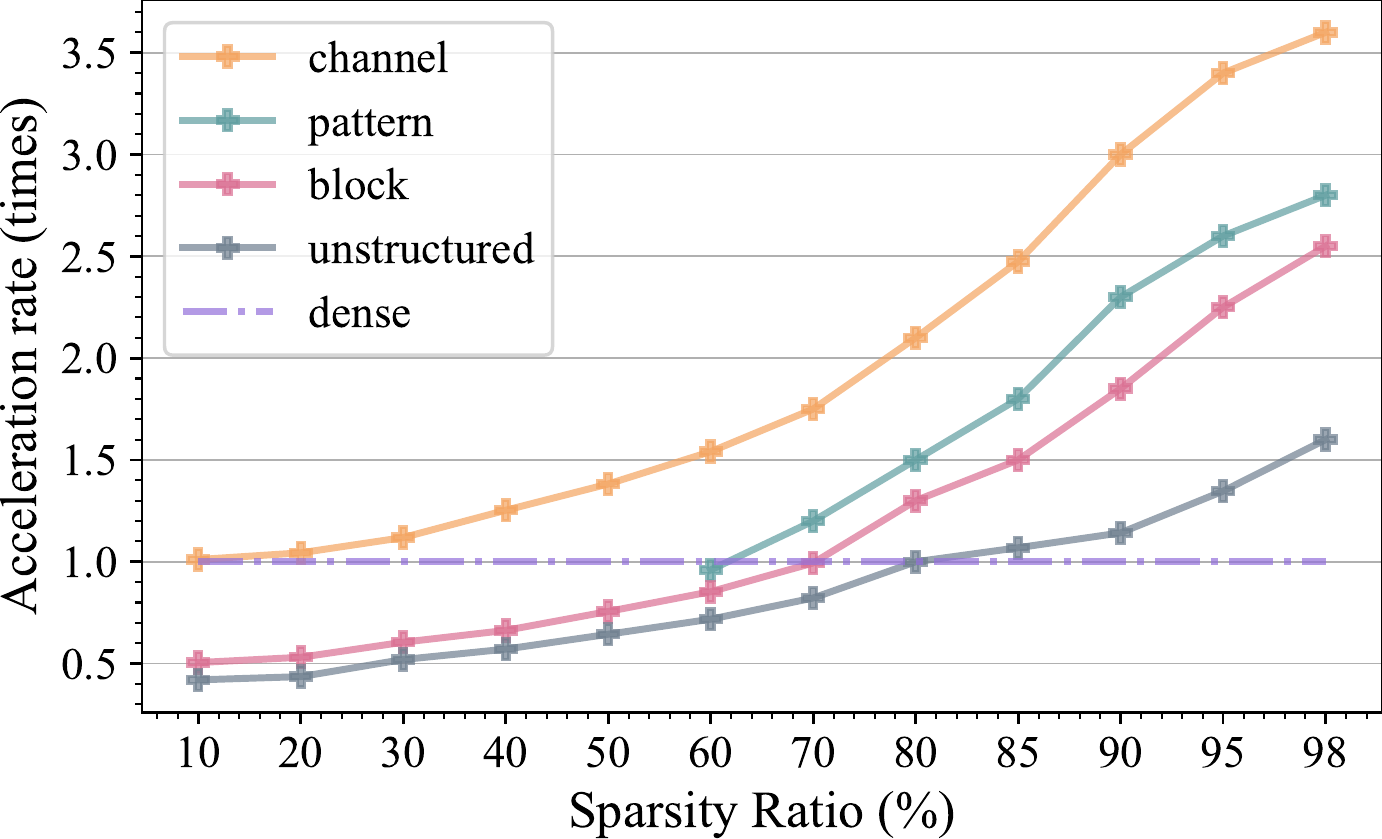}
    \caption{Training acceleration rate vs sparsity ratio of different sparsity schemes. The results here are measured on a 3$\times$3 CONV layer selected from ResNet-32 with 64/64 input/output channels and 16$\times$16 input feature map, with a Samsung Galaxy S20 smartphone. 
    }
    \label{fig:speedup}
\vspace{-0.5em}
\end{wrapfigure}

Figure~\ref{fig:speedup} shows representative training speed performance under various sparsity schemes, using an example CONV layer adopted from ResNet-32.
We evaluate the acceleration rate by measuring the total forward and backward execution time of the sparse CONV layer on a Samsung Galaxy S20 smartphone, and then normalizing with respect to that of the corresponding dense layer. 
Surprisingly, even under the same sparsity ratio, we found that the acceleration rates of different sparsity schemes are varied significantly.
When the sparsity ratio is below 70\% and 80\% for block-based sparsity and unstructured sparsity schemes, respectively, they cannot achieve any acceleration and even slow down the computation speed, compared to the corresponding dense model.
Thus, choosing an appropriate sparsity scheme is an essential factor for sparse training for acceleration purposes.

\subsection{Proposed Memory-Economic Sparse Training (MEST) Framework}
To facilitate the sparse training on edge devices, our MEST framework is designed for the following objectives: 1) towards end-to-end memory-economic training by considering the resource limitation of edge devices; 2) Exploiting sparsity schemes to achieve high sparse training acceleration while maintaining high accuracy.
We propose the MEST method (vanilla) to periodically remove less important non-zero weights from the sparse model and grow zero weights back during the sparse training process, which we call \textbf{mutation}, to explore desired sparse topologies with a specified \textit{sparsity scheme} and \textit{ratio}.
Previous works~\cite{mocanu2018scalable, mostafa2019parameter} directly use the weight magnitude as the indicator of its importance.
However, determining the weight importance only based on its magnitude is not ideal, because the weight magnitude may fluctuate significantly during the training.
Therefore, in MEST, we incorporate the weight's current gradient as an indicator for its changing trend to estimate its importance.
We define the importance score as:
\begin{equation} 
\label{eq:import-est}
Scr_{w_\tau} = |w_\tau |+ |\lambda \frac{\partial \ell(W_{\tau-1}, D)}{\partial w_{\tau-1}}|,
\end{equation}
where $D$ denotes the training data; 
$\ell(W_{\tau-1}, D)$ and $\frac{\partial \ell(W_{\tau-1}, D)}{\partial w_{\tau-1}}$ are the 
loss and gradient at epoch $\tau$; and
$\lambda$ is the coefficient for the gradient.
As a result, three types of weights are considered relatively important, which are the weights with 1) large weight magnitude but small gradient, 2) small weight magnitude but large gradient, and 3) large weight magnitude and large gradient.
The exploration of the impact of the coefficient $\lambda$ on sparse training accuracy is shown in Appendix~\ref{sec:appen_weight_importance}.

\begin{wrapfigure}{R}{0.6\textwidth}
    \begin{minipage}{0.6\textwidth}
      \begin{algorithm}[H]
      \small
      \label{algo:MBST_EM}
        \caption{MEST with (Soft) Elastic Mutation}
          \DontPrintSemicolon
  \KwInput{Network with uninitialized weight $W$ in a total of $L$ layers, target sparsity ratio $s$, $p$, $\tau$, $\Delta \tau$, $\tau_{stop}$.}
  \KwOutput{A sparse model satisfying the target sparsity requirement.}
  Initialize \emph{W} with random values and random sparse mask according to the sparsity requirements. \\
  \While{$\tau < \tau_{stop}$ }
  {     
        \If{MEST+EM}{
            \If{$(\tau \mod \Delta\tau)=0$ \Comment*[r]{do weight mutation}}{
            Decay $p$ if $\tau$ reaches a decaying milestone. \\
            \For{each layer weight tensor $W^{l}$}{ 
                $W^{l} \leftarrow $ \texttt{ArgRemoveTo}$(W^{l},s+p) $ \\
                $W^{l} \leftarrow $ \texttt{ArgGrowTo}$(W^{l},s) $
      		    }
            }
        }
        
        \If{MEST+EM\&S}{
            \ \ Decay $p$ if $\tau$ reaches a decaying milestone. \\
            \For{each layer weight tensor $W^{l}$}{ 
              $W^{l} \leftarrow $ \texttt{ArgGrowTo}$(W^{l},s-p) $ 
  		    }
  		    \ \ Training for $\Delta \tau$ epochs;  \Comment*[r]{$\tau \leftarrow \tau+\Delta \tau$} 
  		    \ \ \For{each layer weight tensor $W^{l}$}{ 
              $W^{l} \leftarrow $ \texttt{ArgRemoveTo}$(W^{l},s) $
  		    }
        }

  }
  Continue sparse training from the epoch $\tau_{stop}$ to $\tau_{end}$.
      \end{algorithm}
    \end{minipage}
\end{wrapfigure}

More importantly, 
different from the methods (e.g., RigL~\cite{evci2020rigging}) that 
use gradients of the dense model to find the weights to grow back, we only use sparse gradients to identify less important weights to remove, then randomly select weights to grow back.
In this way, our MEST strictly keeps the sparsity of weights and gradients during the entire sparse training process. 
This is critical for memory-economic sparse training on resource-limited edge devices.
Moreover, different from previous dynamic sparse training methods that only uses unstructured sparsity, our MEST consider the constraints of different sparsity schemes into the mutation policy (more details in Appendix~\ref{sec:appen_EM_for_schemes}).

\textbf{Elastic Mutation (EM):} We further propose an Elastic Mutation method to gradually reduce the mutation rate along with the training process, called Memory-Economic Sparse Training with Elastic Mutation (MEST+EM).
We are mainly based on two considerations: 1) a larger mutation ratio will provide a larger search space during the dynamic sparse training process; and 2) the dramatic structural change of the network may compromise the training convergence. Thus, we propose our EM method to gradually reduce the mutation ratio during the dynamic sparse training process, which maintains a sufficient search space while making the sparse model smoothly stabilize to an optimized structure. 

\textbf{\emph{Soft memory bound} Elastic Mutation (EM\&S):}
If the application scenario that the memory footprint could be a soft constraint, we propose an enhancement with the \emph{Soft Memory Bound}, namely, Memory-Economic Sparse Training with \emph{Soft-bounded} Elastic Mutation (MEST+EM\&S), as an option to further improve accuracy.
Different from the EM method that the less important weights will always be removed, our EM\&S allows the newly grown weights to be added to the existing weights and then trained, then the less important weights will be selected from all weights including the newly grown weights. This can avoid forcing the existing weights in the model to be removed if they are more important than newly grown weights. This can be considered as adding an ‘undo’ mechanism to the mutation process.
Note that even with a soft memory bound, the target sparsity ratio can still be met by the end of sparse training and keep the entire training process sparse.

\textbf{\textit{Notation and Preliminary:} } Consider $W \in \mathbb{R}^{N}$ is the weights of the entire network. The number of weights in the $l$-th layer $W^{l}$ is $N^l$. Our target sparsity ratio is denoted by  $s \in (0,1)$.
We mutate on a fraction $p \in (0,1)$ of the weights in $W^{l}$. 
Suppose the total number of training epoch is $\tau_{end}$, then we conduct the weight mutation for the first $\tau_{stop} (< \tau_{end})$ epochs with a frequency of $\Delta \tau$. 

Algorithm~\ref{algo:MBST_EM} shows the flow of MEST+EM and MEST+EM\&S.
The main difference between MEST+EM and MEST+EM\&S is the order that $\texttt{ArgRemoveTo}(\cdot)$ and $\texttt{ArgGrowTo}(\cdot)$ are performed. 
In MEST+EM, we perform weight mutation for every $\Delta \tau$ epochs with following steps:
first use $\texttt{ArgRemoveTo}(W^{l},s+p)$ to remove $p\times N^{l}$ less important weights from a total of $s\times N^{l}$ non-zero weights; and then grow the model back to sparsity $s$ with $\texttt{ArgGrowTo}(W^{l},s)$, which randomly activates a number    $p\times N^{l}$ of zero weights. 
The newly activated weights although are being zeros, will be considered as part of the sparse model and be trained. During the entire training process, the model sparsity is strictly bounded by $s$, thus maintaining a hard memory bound.
On the other hand, in MEST+EM\&S, for every other $\Delta \tau$ epochs, we first grow the model to reduce the sparsity ratio to $s-p$ by $\texttt{ArgGrowTo}(W^{l},s-p)$ and train for $\Delta \tau$ epochs, and then remove weights to increase the sparsity ratio back to $s$ by $\texttt{ArgRemoveTo}(W^{l},s)$. We also decay $p$ at given milestone epochs until $\tau_{stop}$. During the entire training process, the weights are trained at sparsity ratio $s-p$, and  sparsity is gradually increasing to the target sparsity ratio through the decay of $p$.
In addition, the mutation process is actually operated on indices, to facilitate implementation on the edge device.

\vspace{-0.5em}
\section{Exploring Data Efficiency in Sparse Training}
\vspace{-0.5em}

Data efficiency has been studied for the traditional training in literature~\cite{bengio2007scaling,kumar2010self,fan2017learning,toneva2018empirical}. 
It has been proven that the amount of information provided by each training example to a network is different, and the difficulty of learning examples also varies.
Some training examples are easily learned at early training stage. And once some examples are learned, they will never be ``forgotten'' (i.e., misclassified) again.
Removing those easy and less informative examples from the training dataset will not cause accuracy degradation on the final model. More details is discussed in Appendix~\ref{sec:appen_dataset}.
However, the exploration of data efficiency in sparse training scenarios is still missing.
Due to the dynamic sparsity mask generation in sparse training, it is still unknown that whether the data efficiency can be leveraged for further accelerating sparse training.
Therefore, we need to first figure out the impact of model sparsity on the number of removable training examples (e.g., unforgettable examples), and then discuss the possibility of leveraging data efficiency for accelerating sparse training.

\vspace{-0.5em}
\subsection{Impact of Model Sparsity on Dataset Efficiency}
\vspace{-0.5em}

In~\cite{toneva2018empirical}, they proposed to use the number of forgetting events of a training example along the entire training process to indicate the amount of information of the example. 
The forgetting event is defined as an individual training example transitions from being classified correctly to incorrectly over the training process.
It can also be considered as the example is forgotten by the network.
An example can be forgotten multiple times. 
An unforgettable example stands for an example that has never been forgotten once it is correctly classified, and it is considered less informative to a network and easy to be learned~\cite{toneva2018empirical}.
More details is shown in Appendix~\ref{sec:appen_dataset}.

\begin{figure*}[t]
    \centering
    \includegraphics[width=0.95\textwidth]{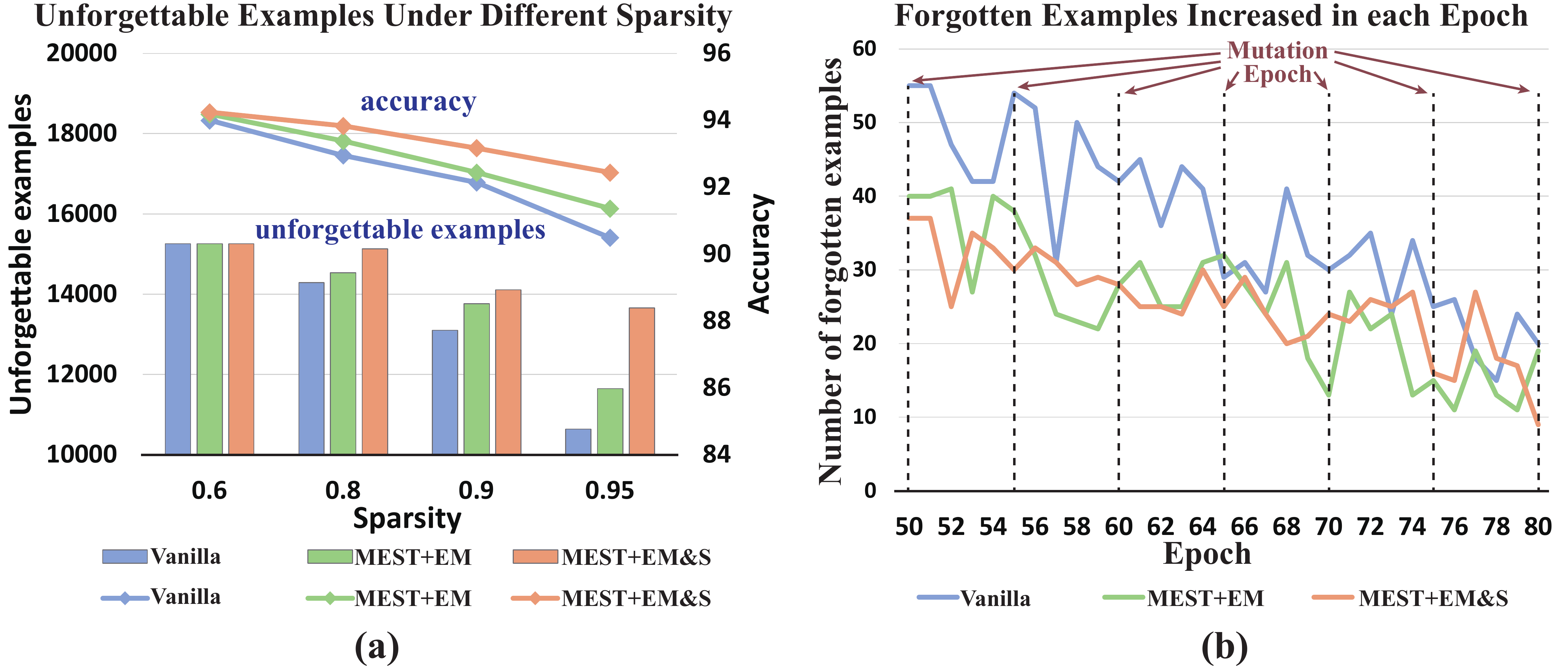}
    \caption{Data efficiency investigation on ResNet-32 using CIFAR-10. (a) The number of unforgettable examples after the sparse training process under different sparse training algorithms and sparsity ratios; 
    (b) The number of increased forgotten examples in each epoch (between epoch 50 to 80 with a mutation frequency of 5 epochs).
    }
    \label{fig:fgt_mut_and_sp}
\end{figure*}

In order to study whether data efficiency can be used to accelerate sparse training, we first explore the number of unforgettable examples that can be identified after the sparse training process under different sparsity ratios.
We test with our three sparse training algorithms MEST (vanilla), MEST+EM, and MEST+EM\&S.
The Figure~\ref{fig:fgt_mut_and_sp} (a) shows the results obtained on ResNet-32 using CIFAR-10 dataset. 
We find that there is still a considerable portion (30\%$\sim$34\%) of training examples in CIFAR-10 dataset are unforgettable to a highly sparse network (under 95\% sparsity).
The number of unforgettable examples decreases as the model sparsity increases, and it shows a positive correlation with the model accuracy.
When under a high sparsity, the network generalization performance decreases, making some easy examples harder to remember.
Moreover, we also observe that a better sparse training algorithm (e.g., MEST+EM\&S) leads to more unforgettable training examples, indicating the potential of removing a larger portion of training examples and hence a higher acceleration.

\vspace{-0.5em}
\subsection{Will Mutation Lead to Forgetting?}
\vspace{-0.5em}
It is a natural question to ask that whether the structure change during the training such as our proposed Elastic Mutation will lead to severe forgetting.
Thus, we evaluate the number of unforgettable examples in the epoch before and after the mutation.
Figure~\ref{fig:fgt_mut_and_sp} (b), shows the number of forgotten examples increased in each training epoch, which equals the difference of unforgettable examples between two consecutive epochs.
Neither the mutation in MEST+EM nor MEST+EM\&S causes a notable increase in forgetting.
This is because the mutated weights are least important, which have a minor impact on the model accuracy.
Detailed results are shown in Appendix~\ref{sec:appen_dataset}.
%

\subsection{Data-Efficient Sparse Training on the Edge}
To identify the less informative training examples, prior work~\cite{toneva2018empirical} collects the statistics of forgetting events through the entire training process.
Then, using compressed dataset to train the network from scratch.
Obviously, this is not an efficient, even an unaffordable solution for training on edge devices.

Different from prior works, we intend to integrate the less informative example identification and data-efficient sparse training into one single training process. Our objective is to obtain a similar final accuracy as a full dataset training within the same number of training epochs. 
Thus, we propose a data-efficient training method (DE), which separate one training process into two phases.
For the first training phase, the full dataset is used for a certain sparse training epochs while counting the number of forgetting events for each example. The first phase takes several epochs (e.g., 70) to obtain a stable identification results.
For the second training phase, partial training examples will be removed from the training dataset and obtain a compressed training dataset for the rest of the training process.
The number of removed examples depends on the number of examples within a forgetting events threshold. 
Note that the examples that only be forgotten few times (e.g., 1 or 2) may also relatively easy to learn, which may also be removed without harming the accuracy.
Denoting the full training dataset as $D$, the compressed training dataset $\hat{D}$ is described as:
\begin{equation} 
\label{eq:dataset}
\hat{D} = \{x_i|x_i \in D \; \text{and} \; f(x_i) \leq th\},
\end{equation}
where the $x_i$ and $f(x_i)$ represent the $i$-$th$ training example in the full training dataset and its number of forgetting events occurred in the first training phase, and $th$ is a given threshold.

\begin{wrapfigure}{R}{0.5\textwidth}
    \centering
    \includegraphics[width=0.5 \textwidth]{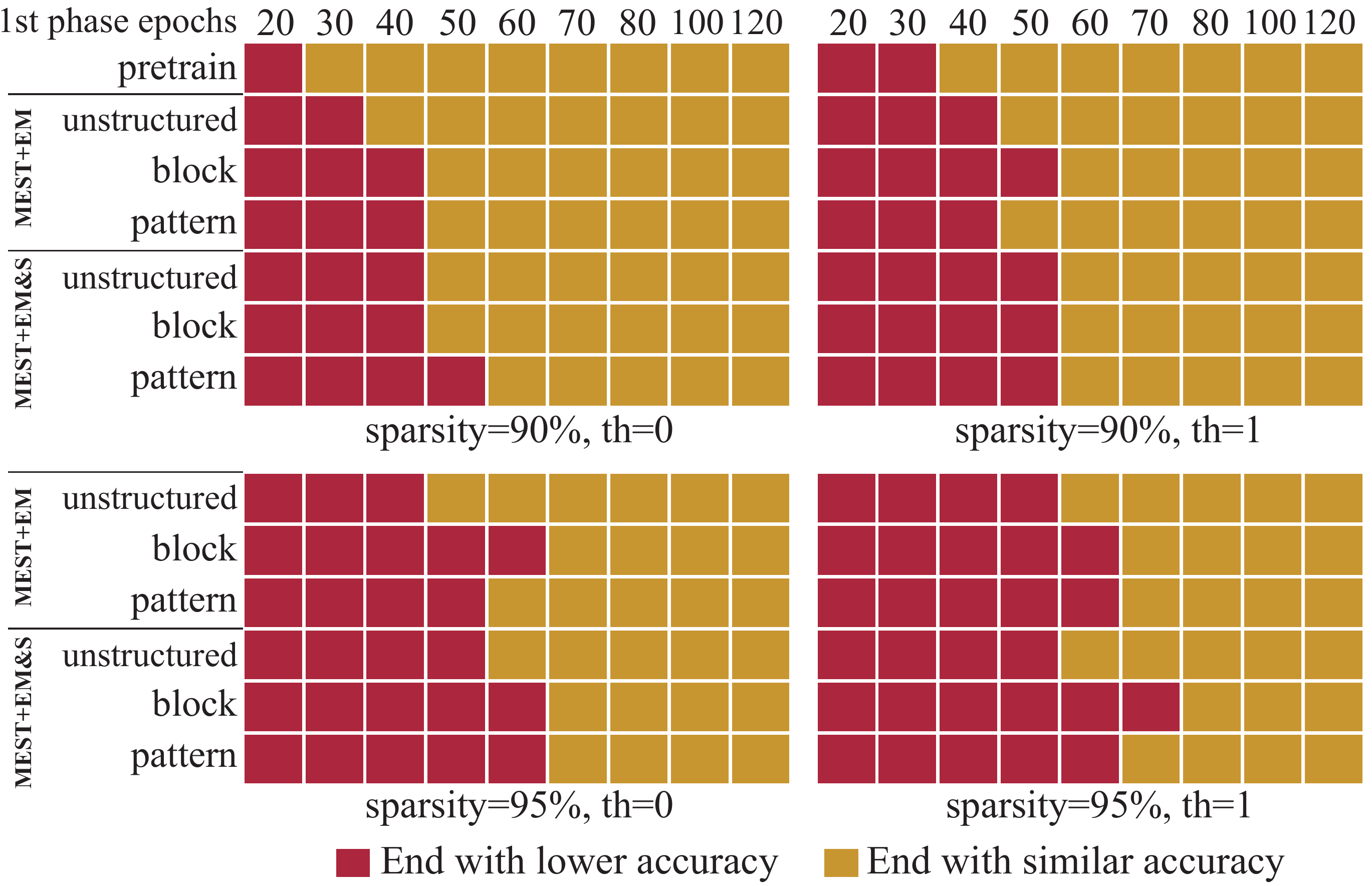}
    \caption{The epoch number used for the first training phase and its corresponding final accuracy status under sparsity ratios of 90\% and 95\% and threshold of 0 and 1. Results are obtained on ResNet-32 using CIFAR-10.
    }
    \label{fig:fgt_epoch}
\end{wrapfigure}

The Figure~\ref{fig:fgt_epoch} shows the status of final accuracy obtained by the two-phase training  approach under different sparsity ratios, sparsity schemes, our proposed mutation methods, number of forgotten thresholds, and epochs for the first phase training.
The yellow grids mean that using that number of epoch for the first training phase can achieve similar accuracy as using a full dataset for the entire training process.

We have the following observations:
1) Compared to dense model, the sparse models takes longer to identify a good set of removable (less informative and easy) examples;
2) The larger number of examples to be removed, the more training epochs are required for the first training phase.
3) Unstructured sparsity scheme requires fewer epochs than fine-grained sparsity schemes (block and pattern).
4) MEST+EM and MEST+EM\&S require similar number of epochs.
5) Besides the unforgettable examples, examples with few forgotten times are also removable without harming the final accuracy.
More results on other dataset and networks can be found in Appendix~\ref{sec:appen_dataset}.
%

\vspace{-0.5em}
\section{Experimental Results}\label{sec:results}
\vspace{-0.5em}

\begin{table*}[t]
\centering
\caption{Accuracy comparison with SOTA works using ResNet-32 on CIFAR-10 and CIFAR-100.}
\vspace{1em}
\scalebox{0.8}{
\begin{tabular}{l | c | ccc | ccc}
\toprule
Dataset    & Memory     & \multicolumn{3}{c}{CIFAR-10}     & \multicolumn{3}{|c}{CIFAR-100}    \\ 
   & Footprint      &
\multicolumn{3}{c}{(\emph{Dense: 94.88})}     & \multicolumn{3}{|c}{(\emph{Dense: 74.94})}    \\
\midrule
Sparsity ratio  &        & 90\%  & 95\%  & 98\%  &  90\%  & 95\%  & 98\%  \\ \midrule
LT~\cite{frankle2018lottery}    & dense  & 92.31 & 91.06 & 88.78     & 68.99 & 65.02 & 57.37 \\ \midrule
SNIP~\cite{lee2018snip}   & dense  & 92.59 & 91.01 & 87.51    & 68.89 & 65.22 & 54.81 \\
GraSP~\cite{wang2019picking}     &  dense   & 92.38 & 91.39 & 88.81    & 69.24 & 66.50  & 58.43 \\ \midrule
DeepR~\cite{bellec2018deep}     &   sparse   & 91.62 & 89.84 & 86.45    & 66.78 & 63.90  & 58.47 \\
SET~\cite{mocanu2018scalable}  & sparse & 92.30  & 90.76 & 88.29     & 69.66 & 67.41 & 62.25 \\
DSR~\cite{mostafa2019parameter} & sparse & 92.97 & 91.61 & 88.46     & 69.63 & 68.20  & 61.24  \\ \midrule
MEST (vanilla)  &  sparse  & 92.12$\pm$0.13 & 90.86$\pm$0.11 & 88.78$\pm$0.26     & 69.35$\pm$0.36 & 67.85$\pm$0.23 & 62.58$\pm$0.31 \\
MEST+EM    &  sparse  & 92.56$\pm$0.07 & 91.15$\pm$0.29 & \textbf{89.22$\pm$0.11}     & \textbf{70.44$\pm$0.26} & \textbf{68.43$\pm$0.32} & \textbf{64.59$\pm$0.27} \\
MEST+EM\&S & sparse & \textbf{93.27$\pm$0.14} & \textbf{92.44$\pm$0.13} & \textbf{90.51$\pm$0.11}   & \textbf{71.30$\pm$0.31} & \textbf{70.36$\pm$0.05} & \textbf{67.16$\pm$0.25} \\ \bottomrule
\end{tabular}}
\label{tab:cifar_results}
\vspace{-0.5em}
\end{table*}

This section  evaluates the  proposed MEST framework. 
The training speed results are obtained using a Samsung Galaxy S20 smartphone with Qualcomm Adreno 650 mobile GPU.
We measure the computation time of a round of forward- and backward-propagation on a batch of 64 images to denote the training speed.
The acceleration rate is the training speed of sparse training normalized to that of dense training.
For accuracy, we repeat each experiment 3 times and report the mean and standard deviation of the accuracy results on CIFAR-10/100.
For training speed, we report the average value from 100 runs.
We use ResNet-32 and VGG-19 for CIFAR-10 and CIFAR-100 dataset~\cite{krizhevsky2009learning}, and ResNet-34 and ResNet-50~\cite{he2016deep} for ImageNet-2012~\cite{deng2009imagenet}.
Since the ImageNet-2012 is not practical for edge training, we mainly use it for accuracy (detailed explanations are in Appendix~\ref{sec:appen_other_network}).
When combining our data-efficient two-phase training method that compresses the training dataset on the second phase, we denote our methods as MEST+EM+DE and MEST+EM\&S+DE.

\textbf{Experimental setups:}
We use the same training epochs as GraSP~\cite{wang2019picking}, which is $\tau_{end}=160$ for CIFAR-10/100 and $\tau_{end}=150$ for ImageNet. 
We use standard data augmentation, and cosine annealing learning rate schedule is used with SGD optimizer. 
For CIFAR, we use a batch size of 64 and set the initial learning rate to 0.1. For ImageNet, we use a batch size of 2048. 
Our learning rate is scheduled with a linear warm-up for 8 epochs before reaching the initial learning rate value of 2.048.
We adopt a uniform sparsity ratio across all the CONV layers while keeping the first layer dense. 
The other reference works (except SET~\cite{mocanu2018scalable} and RigL~\cite{evci2020rigging} that use uniform sparsity) use non-uniform sparsity, which leads to a higher computation FLOPs compared to the uniform sparsity under the same sparsity ratio.
An ablation study of using hybrid sparsity schemes and non-uniform sparsity ratio on MEST is shown in Appendix~\ref{sec:appen_layerwise_scheme_and_ratio}.
The hyper-parameter setting for elastic mutation are provided in Appendix~\ref{sec:appen_setup}.

\subsection{Accuracy Results} \label{sec:accuracy_results}
\textbf{CIFAR-10 and CIFAR-100:} 
The MEST accuracy results are shown in Table~\ref{tab:cifar_results}. We include the results at sparsity ratios of 90\%, 95\%, and 98\% with unstructured sparsity scheme.
Methods that use dynamic sparse training (DeepR, SET, and DSR) achieve slightly better results compared to fixed-mask sparse training.
Compared to MEST (vanilla), which uses a fixed mutation ratio along with the training process, our MEST+EM consistently achieves higher accuracy. 
This proves the effectiveness and importance of our elastic mutation method in sparse training.
And our MEST+EM\&S further improves the accuracy significantly, especially in extremely high sparsity ratio (e.g. 98\%). 
In terms of peack memory footprint, Lottery Ticket (LT), SNIP, and GraSP are equivalent to dense training.
Because the SNIP and GraSP require computing the forward and backward propagation of a dense model to find a desired sparse structure.
The LT method requires an iterative magnitude pruning process to find the ``winning ticket'' sparse structure first, it is also considered the same as the dense model in an end-to-end training scenario~\cite{frankle2018lottery, ma2021sanity,liu2021lottery}.
The VGG-19 results are in Appendix~\ref{sec:appen_other_network}.

\textbf{ImageNet-2012:} 
Table~\ref{tab:imagenet_results} shows the accuracy results and training FLOPS using ResNet-50. 
RigL~\cite{evci2020rigging} is a recent milestone of dynamic sparse training works, which has considerable improvements compared to previous works.
To make a fair comparison with RigL, we scale our training epochs to have the same or less overall training FLOPs as the RigL.
We also increase the training effort for MEST by 1.7$\times$, which is 250 epochs, to compare with the RigL with 5$\times$ longer training, which is 500 epochs as reported in~\cite{evci2020rigging}.
With the same or less training FLOPS, our proposed MEST framework consistently outperforms other baselines.
When using our data effective training method, training FLOPS can be further reduced while maintaining the same accuracy as using the full dataset.
Note that the RigL exploits the dense model gradients to dynamically select the model structure during the sparse training process, which requires frequent dense backpropagations to calculate the dense gradients, and it is not memory-economic for Edge devices.
The more analysis and results for ResNet-34 are shown in Appendix~\ref{sec:appen_other_network}.

\begin{table}[t]
\centering
\caption{Accuracy comparison using ResNet-50 on ImageNet using unstructured sparsity scheme.}
\scalebox{0.82}{
    \begin{tabular}{l | ccc | ccc}
        \toprule
        Method & \multicolumn{1}{c}{Training} & \multicolumn{1}{c}{Inference} & \multicolumn{1}{c}{Top-1} & \multicolumn{1}{|c}{Training} & \multicolumn{1}{c}{Inference} & Top-1    \\
        
        & \multicolumn{1}{c}{FLOPS} & \multicolumn{1}{c}{FLOPS} & \multicolumn{1}{c}{Accuracy} & \multicolumn{1}{|c}{FLOPS} & \multicolumn{1}{c}{FLOPS} & Accuracy \\ 
        
        & \multicolumn{1}{c}{($\times$e18)} & \multicolumn{1}{c}{($\times$e9)} & \multicolumn{1}{c}{(\%)} & \multicolumn{1}{|c}{($\times$e18)} & \multicolumn{1}{c}{($\times$e9)} & (\%) \\ \midrule
        
        Dense & \multicolumn{1}{c}{4.8} & \multicolumn{1}{c}{8.2} & \multicolumn{1}{c}{76.9}  & &  &  \\ \midrule
        Sparsity ratio & \multicolumn{3}{c}{80\%} & \multicolumn{3}{|c}{90\%} \\ \midrule
        SNIP~\cite{lee2018snip} & 1.67  & 2.8 & 69.7   & 0.91 & 1.9  & 62.0    \\
        GraSP~\cite{wang2019picking} & 1.67  & 2.8 & 72.1  & 0.91 & 1.9 & 68.1    \\
        \midrule
        DeepR~\cite{bellec2018deep}  & n/a & n/a & 71.7 & n/a & n/a & 70.2     \\
        SNFS~\cite{dettmers2019sparse} & n/a  & n/a & 73.8   & n/a  & n/a & 72.3     \\
        DSR~\cite{mostafa2019parameter} & 1.28  & 3.3 & 73.3  & 0.96  & 2.5 & 71.6     \\
        SET~\cite{mocanu2018scalable}  & 0.74 & 1.7 & 72.6   & 0.32 & 0.9 & 70.4     \\
        RigL~\cite{evci2020rigging}  & 0.74 & 1.7 & 74.6   & 0.39 & 0.9 & 72.0 \\
        RigL$_{5\times}$~\cite{evci2020rigging}  & 3.65 & 1.7 & 76.6   & 1.95 & 0.9 & 75.7 \\
        \midrule
        
        MEST$_{0.5\times}$+EM\&S & \textbf{0.74}   & 1.7 & \textbf{75.11} & \textbf{0.39} & 0.9  & \textbf{72.37}   \\ 
        MEST$_{0.67\times}$+EM & \textbf{0.74}  & 1.7  & \textbf{75.39}  & \textbf{0.39} &  0.9  & \textbf{72.58}   \\ 
        MEST$_{0.5\times}$+EM\&S+DE & \textbf{0.70}  & 1.7  & \textbf{75.09} & \textbf{0.37} & 0.9   & \textbf{72.36}   \\
        \midrule

        MEST+EM & \textbf{1.10} & 1.7  & \textbf{75.75}   & \textbf{0.48} &  0.9  & \textbf{73.63}   \\
        MEST+EM\&S & \textbf{1.27}  & 1.7  & \textbf{75.73} & \textbf{0.65} & 0.9  & \textbf{75.00}   \\ 
        MEST+EM\&S+DE & \textbf{1.17}  & 1.7  & \textbf{75.70} & \textbf{0.60} &  0.9  & \textbf{75.10}   \\ \midrule
        MEST$_{1.7\times}$+EM & \textbf{1.84}  & 1.7  & \textbf{76.71}  & \textbf{0.80}& 0.9    & \textbf{75.91}   \\
        MEST$_{1.7\times}$+EM\&S & \textbf{2.15} &  1.7   & \textbf{77.19} & \textbf{1.11} & 0.9  & \textbf{76.13}   \\
        MEST$_{1.7\times}$+EM\&S+DE &  \textbf{1.96} & 1.7   & \textbf{77.11} & \textbf{1.01} & 0.9   &  \textbf{76.08}  \\
        \bottomrule
    \end{tabular}
}
\label{tab:imagenet_results}
\end{table}

\textbf{Exploring Sparsity Schemes.}
Figure~\ref{fig:acc_speedup_acc_vs_scheme} (a) and (b) illustrates the accuracy by different sparsity schemes, i.e., unstructured, structured (channel), and fine-grained structured (block and pattern) when using our MEST+EM and MEST+EM\&S, respectively.
In block-based sparsity scheme, we set the block size as $(4,1)$. And in pattern-based sparsity scheme, we use 8 sparse patterns according to \cite{ma2020image}.

We evaluate MEST framework with a sparsity ratio ranging from 10\% to 98\%. 
Note that due to structural nature of pattern-based pruning, its sparsity ratio must be at least 55.6\% (see Appendix~\ref{sec:appen_scheme}).
To make a fair comparison, we only choose the reference works that can maintain both sparse weights and gradients along the entire training process.
Figure~\ref{fig:acc_speedup_acc_vs_scheme} (a) and (b) shows that all other sparsity schemes outperform channel sparsity scheme as expected, but the accuracy of channel sparsity scheme can be improved with weight elastic mutation. 
Our MEST+EM results show that the accuracy of our unstructured scheme results are higher than reference works and our block-based and pattern-based scheme results under all sparsity. The block-based and pattern-based schemes achieve similar accuracy as our unstructured scheme when sparsity is lower than 80\%.
With our MEST+EM\&S, the accuracy of all sparsity schemes are boosted and outperforms the reference works.

\subsection{Memory Footprint and Training Acceleration by Sparsity and Data-Efficient Training} \label{sec:acceleration}

\begin{figure}[t!]
    \centering
    \includegraphics[width=1 \columnwidth]{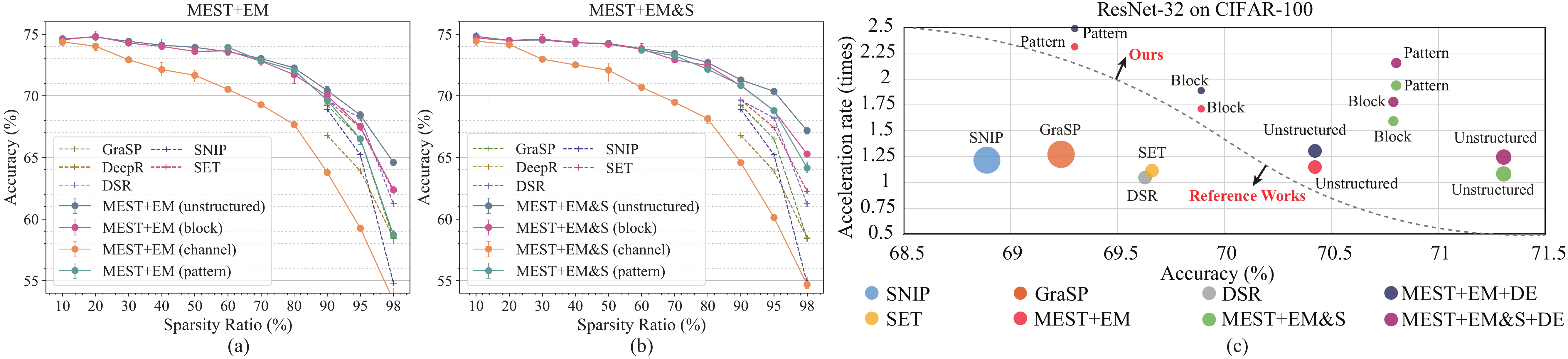}
    \caption{Results obtained using ResNet-32 on CIFAR-100. We choose the reference works that can maintain both sparse weights and gradients along the entire training process. (a) and (b) Accuracy results of the proposed MEST framework using different sparsity schemes and sparsity. (c) Comparison with representative SOTA works in accuracy, training acceleration rate, and memory footprint. Sparsity ratio is 90\% for all results. The acceleration rate is normalized with respect to dense training. The size of the circles represents the relative cost of the memory footprint. We use $th=0$ as the threshold for DE.}
    \label{fig:acc_speedup_acc_vs_scheme}
\end{figure}

In Figure~\ref{fig:acc_speedup_acc_vs_scheme} (c), we compare the model accuracy, training acceleration, and memory footprint among our MEST and representative SOTA works, i.e.,
SET, DSR, SNIP, and GraSP.
We show the results using ResNet-32 on CIFAR-100 with 90\% sparsity.
The training acceleration results are normalized to dense training. 
Note that the SNIP and GraSP require computing the forward and backward propagation of a dense model to find a desired sparse structure. Therefore, their memory footprint is considered the same as the dense model in an end-to-end training scenario.
The area of the circles represents the relative costs of the memory footprint (the smaller the better).

With unstructured sparsity, all reference works and our unstructured sparsity (without DE) results can only achieve minor training acceleration (1.04$\times\sim1.27\times$), even under such a high sparsity ratio.
Because the unstructured sparsity leads to irregular memory access which introduces significant execution overhead.
Moreover, the dense model can take advantage of Winograd~\cite{lavin2016fast} to significantly accelerate the computation speed, but cannot be applied to sparse model. 
For our block-based and pattern-based sparsity schemes, without applying our data-efficient training, the acceleration rates are greatly increased and achieve up to 2.3$\times$ acceleration compared to the dense training while maintaining comparable accuracy.
Our data-efficient training approach can effectively provide an extra speedup to all our results. 
The speedup is from 10\% to 15\% while not compromising the accuracy. The acceleration from data-efficient training is much higher on the CIFAR-10 dataset (20.6\%$\sim$22.5\%) since more examples are unforgettable and can be removed (more details in Appendix~\ref{sec:appen_other_network}).
Comparing with SNIP, GraSP, SET, and DSR, the best-performant MEST increases accuracy by 1.91\%, 1.54\%, 1.14\%, and 1.17\%; 
achieves 1.76$\times$, 1.65$\times$, 1.87$\times$, and 1.98$\times$ training acceleration rate; 
and reduces the memory footprint by 8.4$\times$, 8.4$\times$, 1.2$\times$, and 1.2$\times$, respectively.

From the memory footprint perspective,  SNIP and GraSP involve dense model during pruning at initialization, their memory footprints are considered the same as the dense model.
Since the fewer indices are needed, the block-based sparsity and pattern-based sparsity under both MEST+EM and MEST+EM\&S methods achieve a smaller memory footprint than all reference works and our unstructured sparsity scheme. More detailed discussion is in Appendix~\ref{sec:appen_memory}..
And a discussion about why it is critical to be memory-economic in edge training is shown in Appendix~\ref{sec:appen_why_mem_eco_critical}.

Our results show a clearer advantage compared to the reference works. 
Even without DE acceleration, when using MEST+EM with block-based or pattern-based sparsity, or using MEST+EM\&S with block-based sparsity, our results still outperform all reference works in all accuracy, training acceleration, and memory footprint aspects.

\textbf{Discussion.}
The pattern-based sparsity shows a consistently better performance in acceleration than block-based sparsity.
However, the accuracy comparison between pattern-based sparsity and block-based sparsity is varied according to the sparsity ratio and dataset. 
Since the pattern-based sparsity is only applicable to 3$\times$3 CONV layers, for the network with different types of layers, a layer-wise sparsity scheme assignment is desired. 
Investigations of  hybrid sparsity schemes are provided in Appendix~\ref{sec:appen_layerwise_scheme_and_ratio}.

On the other hand, since both dataset compression and model sparsity make the trade-off between accuracy and acceleration, it is interesting to investigate the performance of different combinations of these two methods (more details in Appendix~\ref{sec:appen_ablation_combination}).

\vspace{-0.5em}
\section{Conclusion}\label{sec:conclusion}
\vspace{-0.5em}
This paper proposes a Memory-Economic Sparse Training (MEST) framework with enhancements by Elastic Mutation and Soft Memory Bound Elastic Mutation.
Then, this paper systematically investigates the sparse training problem with respect to the sparsity schemes.
We implement a prototype design on a mobile device to accurately measure the training speed performance.
We investigate and incorporate the data-efficient training in sparse training scenario to further boost the acceleration.
With MEST framework, a feasible solution is provided for sparse training on edge devices with superior performance on accuracy, speed, and memory footprint.

\section*{Acknowledgment}
This work is partly supported by the National Science Foundation CCF-1937500, CCF-1733701, and CCF-2047516 (CAREER), Army Research Office/Army Research Laboratory via grant W911NF-20-1-0167 (YIP) to Northeastern University, and Jeffress Trust Awards in Interdisciplinary Research to William \& Mary. Any opinions, findings, and conclusions or recommendations expressed in this material are those of the authors and do not necessarily reflect the views of NSF, ARO, or Thomas F. and Kate Miller Jeffress Memorial Trust.

{
\small
\bibliography{ref}
\bibliographystyle{unsrt}
}

\appendix
\counterwithin{figure}{section}
\counterwithin{table}{section}


\clearpage

{\huge \bf Appendix}

\section{Sparsity Scheme}\label{sec:appen_scheme}

\begin{figure}[h]
    \centering
    \includegraphics[width=0.6 \textwidth]{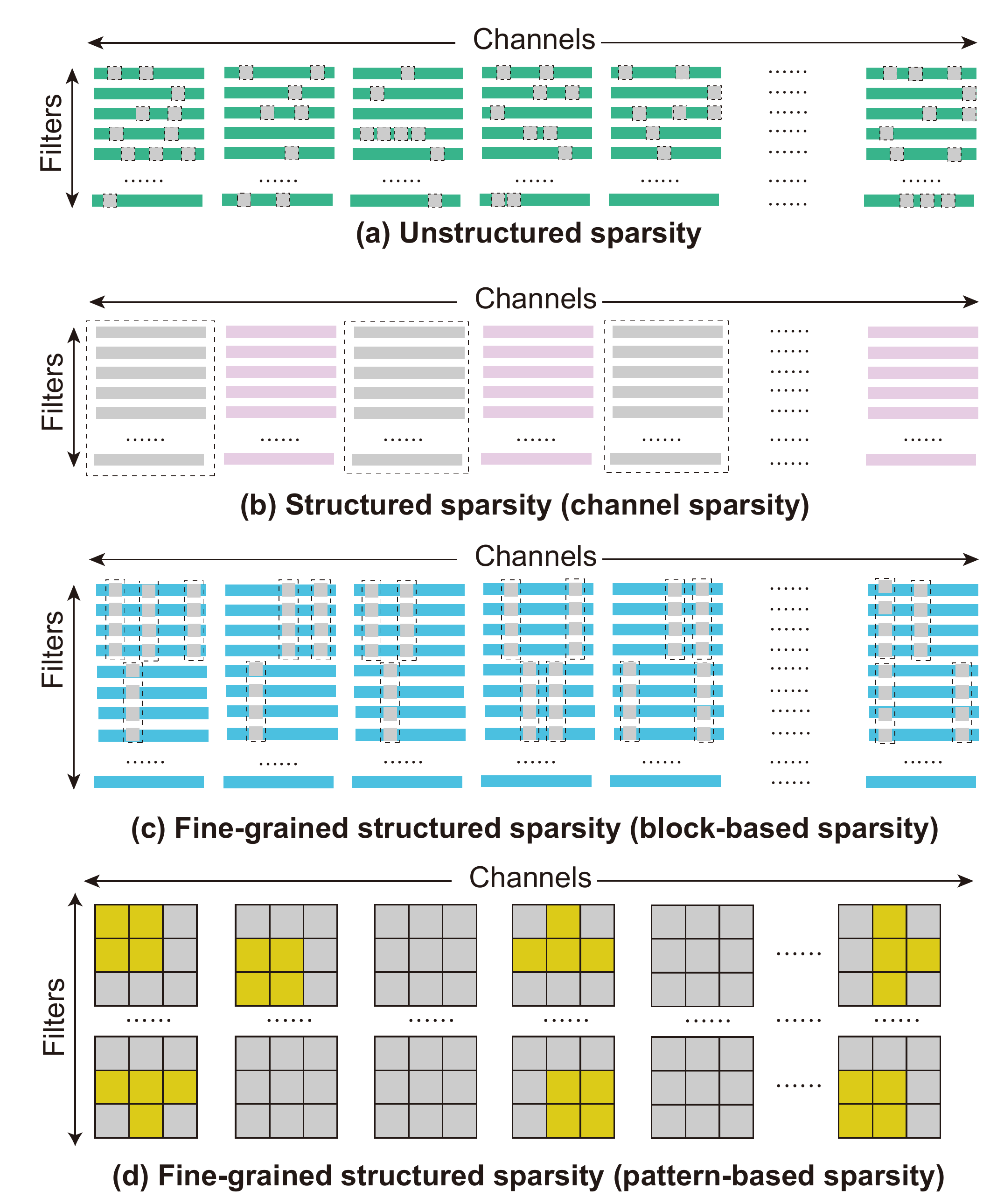}
    \caption{(a) Unstructured sparsity; (b) structured sparsity (channel); (c) fine-grained structured sparsity (block); and (d) fine-grained structured sparsity (pattern).}
    \label{fig:sparsity_schemes}
\end{figure}

We use Figure~\ref{fig:sparsity_schemes} to illustrate existing weight sparsity schemes, where grey represents the zero weights and colored parts are for remaining non-zero weights in a sparse model.
In Figure~\ref{fig:sparsity_schemes} (a)$\sim$(c),  the GEMM matrix format is used for demonstrating the sparsity schemes.
Figure~\ref{fig:sparsity_schemes} (d) illustrates directly on the weight tensor.

Figure~\ref{fig:sparsity_schemes} (a) is the \textbf{unstructured sparsity}~\cite{han2015deep,guo2016dynamic}, 
where zero weights are distributed at arbitrary locations. The unstructured sparsity can achieve a high sparsity ratio with negligible effect on accuracy, but is not compatible with data-parallel executions on computing devices. 

Figure \ref{fig:sparsity_schemes} (b) shows a type of \textbf{structured sparsity}~\cite{luo2017thinet,yu2018nisp,dong2019network,hu2016network,molchanov2017variational,wen2016learning,he2017channel,he2018soft,li2019compressing,he2019filter,neklyudov2017structured,dai2018compressing}, 
called channel sparsity, where
the weights of entire channels are set to  zeros. 
The other type of structured sparsity is the filter sparsity \cite{luo2017thinet,he2018soft,he2019filter,li2019compressing}.
These two types of structured sparsity are somewhat equivalent, 
because if some filters are removed in one layer, the corresponding channels of the next layer become invalid.
The structured sparsity preserves regularity on the sparse models, but suffers from significant accuracy loss.

Figure \ref{fig:sparsity_schemes} (c) and (d) show two types of fine-grained structured sparsity~\cite{yang2018efficient,ma2020pconv,niu2020patdnn,ma2020image,dong2020rtmobile}: block-based sparsity and pattern-based sparsity.

For \textbf{block-based sparsity}, weights are partitioned into blocks with the same size. Figure \ref{fig:sparsity_schemes} (c) illustrates an example block size of $(4,1)$.
In the block-based sparsity, weights in the same block are either set to zeros or remaining together.
Since block-based sparsity uses a much finer granularity compared with structured sparsity, it is considered as a fine-grained structured sparsity.

Figure \ref{fig:sparsity_schemes} (d) shows the \textbf{pattern-based sparsity}, which is a combination of \emph{kernel pattern sparsity} and \emph{connectivity sparsity}.
In kernel pattern sparsity, for each kernel in a filter, a fixed number of weights are set to zeros, and the remaining weights form specific kernel patterns. 
The example in Figure \ref{fig:sparsity_schemes} (d) is defined as 4-entry kernel pattern, since every kernel preserves 4 non-zero weights out of the original 3$\times$3 kernels.
Besides that, the connectivity sparsity cuts the connections between some input and output channels, which is equivalent to removing corresponding whole kernels.

\begin{table*}[h]
\centering
\caption{The impacts of sparsity schemes on the memory footprint in sparse training.}\label{tab:memfootprinteqns}
\scalebox{0.8}{
\begin{tabular}{l|l|l}
\toprule
Scheme & Memory Footprint & Approximation \\ \midrule
Dense & $2N\cdot b_{w}$ & $2N\cdot b_{w}$\\ \midrule
Structured & $(1-s)\cdot(2N\cdot b_{w})$ & $(1-s)\cdot(2N\cdot b_{w})$\\ \midrule
Unstructured & $(1-s)\cdot (2N\cdot b_{w}+N\cdot b_{index})+\sum_l((F_l+1)\cdot b_{index})$ & $(1-s)\cdot(2N\cdot b_{w}+N\cdot b_{index})$\\ \midrule
Pattern & $(1-s)\cdot(2N\cdot b_{w}+\frac{1}{4}N\cdot b_{index})+\sum_l((F_l+1)\cdot b_{index})$ & $(1-s)\cdot(2N\cdot b_{w}+\frac{1}{4}N\cdot b_{index})$ \\ \midrule
Block & $(1-s)\cdot(2N\cdot b_{w}+\frac{1}{B}N\cdot b_{index})+\sum_l((\frac{1}{m}F_l+1)\cdot b_{index})$ & $(1-s)\cdot(2N\cdot b_{w}+\frac{1}{B}N\cdot b_{index})$ \\ \bottomrule
\end{tabular}}
\end{table*}

\section{Memory Footprint}\label{sec:appen_memory}
Consider a sparse model with a sparsity ratio $s \in [0,1]$ obtained from a dense model with a total of $N$ weights.
A higher value of $s$ denotes fewer non-zero weights in the sparse model.
Suppose weights are represented as $b_w$-bit numbers.
Each gradient is therefore represented with $b_w$ bits.
For sparse models, we need indices for denoting the sparse topology of weights/gradients within the dense model. 
Indices are represented as $b_{index}$-bit numbers.
Generally, mobile edge devices can support 8-bit fixed-point, 16-bit floating-point, and 32-bit floating-point numbers.
Weights and gradients are usually using 16-bit or 32-bit.
Due to the data storage format on edge devices, 8-bit or 16-bit is preferred for indices.

Table \ref{tab:memfootprinteqns} lists the memory footprint of sparse training in relevance to the sparsity scheme.
In the \textbf{structured sparsity} scheme, where entire filters/channels are zeros, the sparse model can be reconstructed into a smaller dense model without indices. Thus, the memory footprint is determined by the number of non-zero weights plus the number of corresponding gradients i.e., $(1-s)\cdot 2N$.

For \textbf{unstructured sparsity} scheme, each non-zero weight and its gradient require indices to denote the corresponding location within the dense matrix.
Compressed sparse row (CSR) format is commonly used for sparse storage.
More specifically, consider the weights of a $l$-th CONV layer  reshaped from 4-D tensor to a 2-D weight matrix, where each row represents the weights from a filter. We use $F_l$, $Ch_l$, and $K_l$ to denote  the number of filters (output channels), number of channels (input channels), and kernel size, respectively.
Thus, there are $F_l$ rows and $Ch_l\cdot K_l^2$ columns in a weight matrix. 
In CSR format, each non-zero weight requires a column index i.e., $col\_index$  to denote its  location within the column.
The number of column indices is equal to the number of non-zero weights.
Also, the number of non-zero weights in each row (filter) should be denoted by $row\_index$, which is a vector with $F_l+1$ elements.
The difference between two adjacent elements in $row\_index$ denotes the number of non-zero weights in each row.
Thus, the total number of indices of the entire network for the unstructured sparsity scheme is $(1-s)\cdot N+\sum_l(F_l+1)$. 
And the memory footprint of model representation together with gradients for unstructured sparsity is shown in Table~\ref{tab:memfootprinteqns}. 
Note that the number of filters (i.e., $F_l$) is much smaller than the total number of weights, and therefore the last term can be ignored as an approximate.

For \textbf{pattern-based sparsity} scheme, 
each kernel with non-zero weights requires a $kernel\_index$ to represent the kernel location in a filter. 
For the case of 4-entry kernel pattern used in Figure \ref{fig:sparsity_schemes} (d), it is equivalent to every 4 non-zero weights sharing a $kernel\_index$. 
Similar as the unstructured sparsity, the $row\_index$ is also needed.

For \textbf{block-based sparsity} scheme,
consider the block size of $B=m\times n$, since all the weights within a block will be either zero or non-zero together, it is equivalent to every $B$ non-zero weights sharing a $block\_index$.
As for the $row\_index$ vector, a total of $\frac{1}{m}F_l+1$ elements are needed.

For the pruning-at-initialization algorithms, SNIP and GraSP involve several dense training iterations to determine the importance of initial weights. When under the end-to-end edge training scenario, their peak memory footprint is considered the same as the dense training:
$2N\cdot b_{w}$.

On the other hand, the sparse training work such as SNFS~\cite{dettmers2019sparse} and RigL~\cite{evci2020rigging}, which requires the sparse weights (unstructured) and dense gradients during the sparse training to evaluate the weight importance, the memory footprint is in between dense training and full sparse training and can be approximated as: $(2-s)\cdot N\cdot b_{w}+(1-s)\cdot N\cdot b_{index}$.

\section{Compiler-Level Optimizations for Training Acceleration}\label{sec:appen_compiler}

\subsection{How Does Sparsity Accelerate Training?} 
\label{sec:sparsityhelpstraining}

The training process consists of two phases, the \textit{forward propagation} and the \textit{backward propagation}.
Considering the $l$-th convolutional (CONV) layer in the neural network, the forward propagation phase during training, which is the same as the inference process, can be formulated as:
\begin{equation}
\label{equ:1}
a^l=\sigma\left(z^l\right)=\sigma\left(W^{l} * a^{l-1}+b^{l}\right),
\end{equation}
where $W$, $b$, and $z$ represent the weights, biases, and output before activation, respectively; $\sigma(\cdot)$ denotes the activation function;  $a$ is the activations; $*$ means convolution operation. 
Many previous works~\cite{he2018soft,he2019filter,li2019compressing,ma2020pconv,niu2020patdnn,dong2020rtmobile} have proved that the sparse weight matrices (tensors) can result in inference acceleration by effectively reducing the number of multiplications in convolution operation.
Thus, the sparsity can inherently accelerate the forward propagation phase in the training process.

On the other hand, the goal of the backward propagation phase is to obtain the gradients of the weights, so as to update the weights.
The two main calculation steps  are as follows:
\begin{align}
& \delta^{l}=\delta^{l+1} * \operatorname{rotate} 180^\circ\left(W^{l+1}\right) \odot \sigma^{\prime}\left(z^{l}\right), \label{equ:2} \\
& G^{l}= a^{l-1} * \delta^{l}, \label{equ:3}
\end{align}
where $\delta^{l}$ is the error associated with the $l$-th layer; and $G^{l}$ denotes the gradients.
In the above equations, $\odot$ represents element-wise product,  $\sigma^{\prime}(\cdot)$ denotes the derivative of activation, and $\operatorname{rotate} 180^\circ(\cdot)$ means rotating matrix 180 degrees.

It can be observed that the computations in both two steps are essentially based on convolution (i.e., matrix multiplication). The former uses sparse weight matrix (tensor) as the operand, and therefore can be accelerated in the same way as the forward propagation.
The latter allows a sparse output result since the gradients have the same sparsity topology as the weights.
Thus, both two steps have reduced computations, which are roughly proportional to the sparsity ratio, and therefore can be accelerated in the back propagation.

\subsection{Compiler Optimizations}
In previous works such as PatDNN~\cite{niu2020patdnn}, compiler-level optimizations are used for accelerating inference.
We extend those compiler optimizations and incorporate various sparsity schemes  for accelerating the sparse training computation in both the forward-propagation and backward-propagation on edge devices.
We adopt several compiler optimization techniques, including \textit{sparse model storage}, \textit{matrix reorder}, and \textit{parameter auto-tuning} to relieve the poor memory performance, heavy control-flow instructions, thread divergence, and load imbalance caused by sparse computation, and thus achieving the sparse training acceleration.

\textbf{Sparse Model Storage.}
Based on the CSR format for unstructured sparse model representation, we use more compact model storage formats delicately designed for pattern-based sparsity and block-based sparsity, which can better compress the storage for indices by leverage the structural regularity of the sparsity schemes and save memory-bandwidth of edge devices.
Moreover, the data locality is further improved, enabling later branch-less execution.

\textbf{Matrix Reorder.}
For the computation during the forward- and backward- propagation for each layer, the matrix multiplication is executed by multiple GPU threads simultaneously. 
Since the weight matrix is highly sparsified, and the non-zero weights are not evenly distributed across the whole weight matrix, the threads may execute the patterns/blocks with significantly divergent computations if the computation follows the original matrix order.
Thus, we introduce the matrix reorder optimization to group the rows (filters) in the weight matrix that have similar computation patterns together (i.e., grouping the rows containing a similar number of non-zero patterns/blocks  to be computed.)
After reordering the matrix, the rows in each group are assigned to multiple threads to achieve balanced processing.

\textbf{Parameter Auto-tuning.}
Sparse training on edge devices involves many execution-related and performance-critical tunable parameters such as the memory data placement, matrix tiling sizes, looping unrolling factors, etc. These parameters will significantly vary the computation efficiency as well as the training speed.
The best-suited configuration of the parameters is hard to be determined manually. 
Thus, We introduce the parameter auto-tuning technique to search the parameters in an automatic manner.

\section{Elastic Mutation for Pattern-based  and Block-based Sparsity Schemes}
\label{sec:appen_EM_for_schemes}

Unlike the unstructured sparsity scheme, the mutation processes of pattern-based and block-based sparsity schemes are required to satisfy the structural constraints of those sparsity schemes.

\textbf{Pattern-based sparsity: } We perform \texttt{ArgRemoveTo($\cdot$)} by removing the least important convolution kernels to meet the sparsity ratio setting. Note that the importance of a specific convolution kernel can be obtained by summing up weight importance in the kernel. For \texttt{ArgGrowTo($\cdot$)}, we randomly select empty kernels (i.e., all weights in kernel are zeros) and set a random pattern style to them. The newly activated weights can be trained from their initial values, which are zeros. The total activated weights should meet the \texttt{ArgGrowTo($\cdot$)} sparsity setting. Please note that \texttt{ArgRemoveTo($\cdot$)} and \texttt{ArgGrowTo($\cdot$)} select the same number of kernels to remove or grow in a convolution filter within each layer for a balanced computation regime~\cite{ma2020pconv,ma2020image}.

\textbf{Block-based sparsity: } We perform \texttt{ArgRemoveTo($\cdot$)} by removing the least important blocks to meet the sparsity ratio setting. Note that the importance of a specific block can be obtained by summing up weigh importance  in the block. For \texttt{ArgGrowTo($\cdot$)}, we randomly select empty blocks (i.e., all weights in block are zeros) and grow the whole block from zero values. The total activated weights should meet the \texttt{ArgGrowTo($\cdot$)} sparsity setting.

\section{Weight Importance Estimation}
\label{sec:appen_weight_importance}

In MEST+EM(\&S), besides the weight magnitude, we also consider the weight's current gradient as an indicator for its changing trend to estimate its importance.
Because the weight with relatively large magnitude may become smaller, indicating it is becoming unimportant, while the small-magnitude weights can become larger as well.
According to the Equation (\ref{eq:import-est}) in the main paper, we consider three types of weights are relatively important in our mutation process, which are the weights with 1) large weight magnitude but small gradient, 2) small weight magnitude but large gradient, and 3) large weight magnitude and large gradient.

\begin{figure}[h]
    \centering
    \includegraphics[width=0.45 \textwidth]{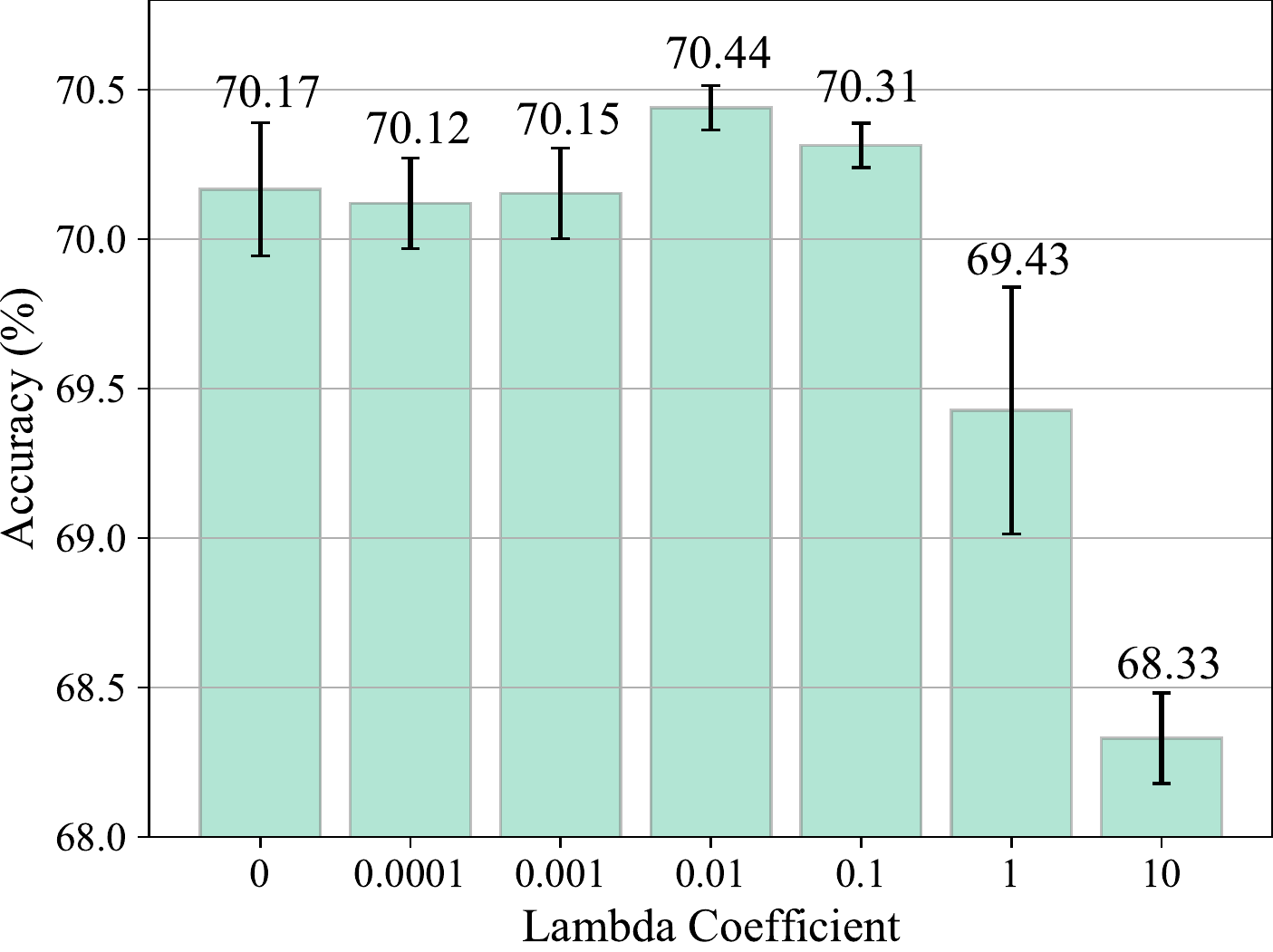}
    \caption{MEST+EM accuracy with varying  $\lambda$ coefficient using ResNet-32 on CIFAR-100.}
    \label{fig:ablation_lambda}
\end{figure}

Figure~\ref{fig:ablation_lambda} shows the sparse trained model accuracy under different $\lambda$ values. We use ResNet-32 on CIFAR-100 as an example.
When $\lambda$ is set to 0.01 or 0.1, the accuracy can be improved compared to only considering the weight magnitude (i.e., $\lambda=0$).

\section{Data-Efficient Training}
\label{sec:appen_dataset}

\subsection{Basic Concepts}
To explore the data efficiency in DNN training, measuring the effective information of a training sample for a network is a very important aspect.
In~\cite{toneva2018empirical}, the number of forgetting events of a training example during the training process is used as an indicator to reflect the amount of information and the complexity of an example.

\textbf{Learning event.} A learning event occurs when a training sample goes from being misclassified to being correctly classified by a network in two consecutive training epochs.

\textbf{Forgetting event.} A Forgetting event occurs when a training sample goes from being correctly classified to being misclassified by a network in two consecutive training epochs.

\textbf{Unforgettable example.} Throughout the entire training process, if an example will never be misclassified after it has been correctly classified, the example is considered an unforgettable example. 
The examples that have never been correctly classified are not considered to be unforgettable.

According to prior works~\cite{bengio2007scaling,kumar2010self,fan2017learning,toneva2018empirical}, the unforgettable examples are generally considered as less informative and easy to be learned.
The figure shows a example of unforgettable training examples and training examples with the highest forgetting event counts obtained using ResNet-32 on CIFAR-10 dataset. 
It can be observed that the unforgettable examples are intuitively much easier to recognize, which preserve distinctive object features and the objects have high contrast to the background. 
On the contrary, the most forgettable examples are clearly more complex compared to the unforgettable examples.

\begin{figure}[h]
    \centering
    \includegraphics[width=1 \textwidth]{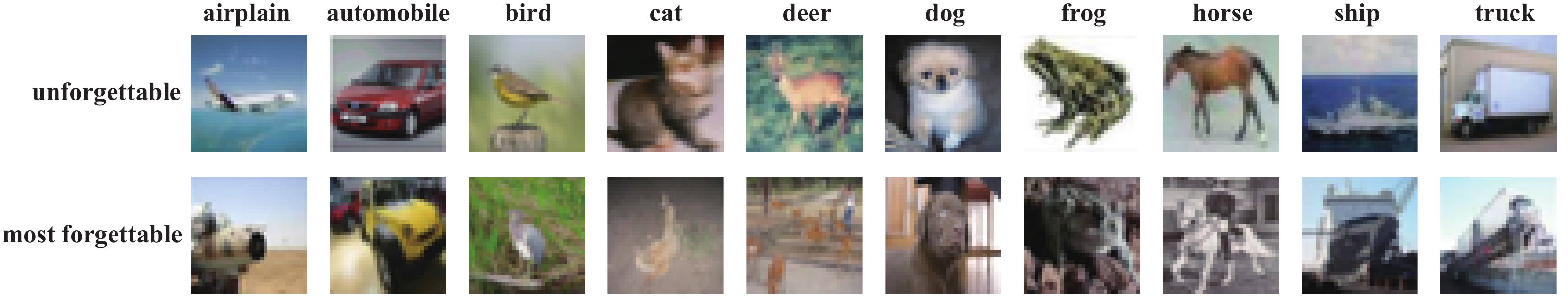}
    \caption{Visualization of unforgettable examples and the most forgettable examples obtained using ResNet-32 on CIFAR-10.}
    \label{fig:visual}
\end{figure}

\subsection{More Results for Final Accuracy using Different Number of Phase-1 Epochs}
\begin{figure}[h]
    \centering
    \includegraphics[width=1 \textwidth]{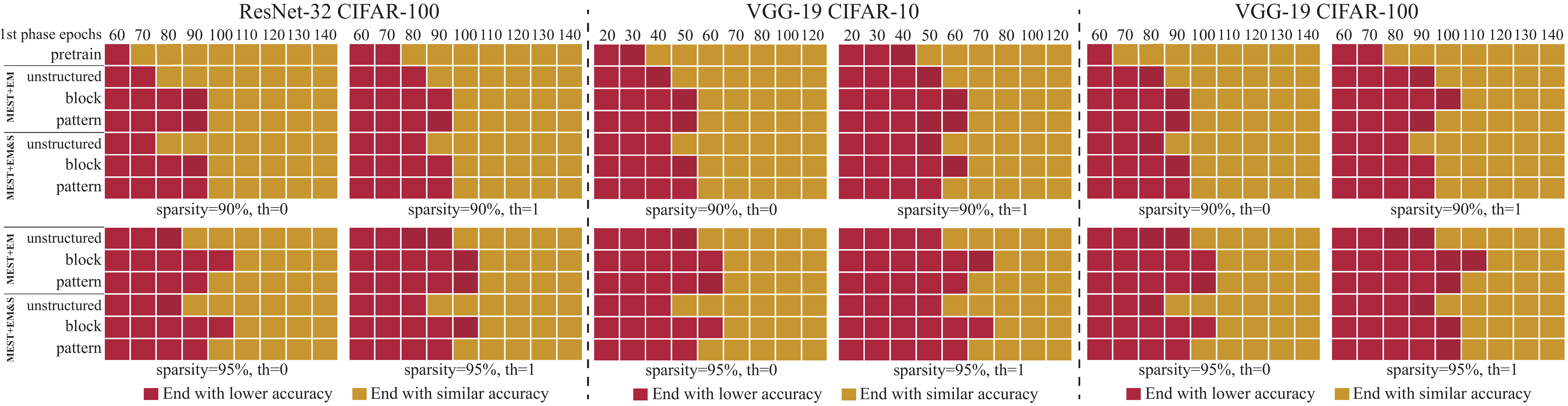}
    \caption{The epoch number used for the first training phase and its corresponding final accuracy status under sparsity ratios of 90\% and 95\% and threshold of 0 and 1.}
    \label{fig:appen_grid}
\end{figure}

Figure~\ref{fig:appen_grid} shows the final accuracy status obtained using different number of first training phase epochs.
The results include ResNet-32 on CIFAR-100, VGG-19 on CIFAR-10, and VGG-19 on CIFAR-100.
The yellow grids stand for using that number of epoch for the first training phase can achieve similar accuracy as using a full dataset for the entire training process.
Compared to the results on the CIFAR-10 dataset, the networks generally require more first training phase epochs on the CIFAR-100 to achieve similar accuracy as the full dataset training. 
This phenomenon can be observed for all pretraining and sparse training methods.

\subsection{Impact of Mutation on Unforgettable Examples}
\begin{figure}[h]
    \centering
    \includegraphics[width=0.75 \textwidth]{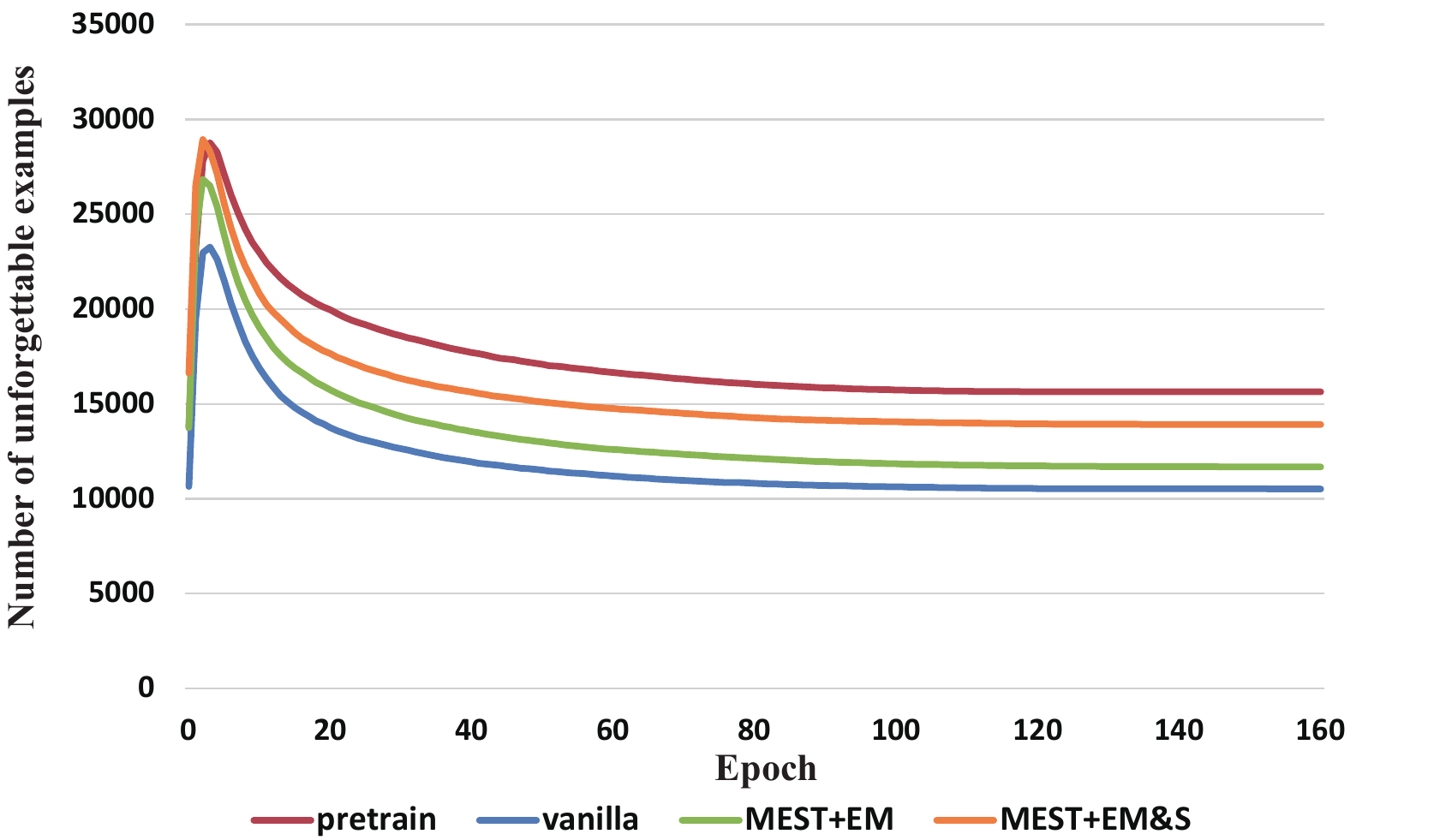}
    \caption{The trend of the number of unforgettable examples throughout the entire training process.}
    \label{fig:appen_trend}
\end{figure}

Figure~\ref{fig:appen_trend} shows the trend of the number of unforgettable examples throughout the entire training process, and we name it the forgetting curve.
Note that different from the MEST(vanilla) method, the vanilla method here stands for a static sparse training method, which randomly pruned weights at initialization and without any mutation along the training process.
Compared to the methods without mutation (i.e., pretrain and vanilla), the forgetting curve of the mutation methods (i.e., MEST+EM and MEST+EM\&S) do not show severe fluctuations throughout the entire training process, indicating that our mutation method will not cause a notable increase in forgetting.
This is because the mutated weights are least important, which only have a minor impact on the model performance. And our elastic mutation also gradually decreases the mutation rate, which further enhance the network stability.
We can also observe that our MEST methods can increase the number of unforgettable examples compared to the vanilla method, which provides the potential of removing a larger portion of training examples and hence a higher acceleration.

\newpage

\section{Experiment Setup}
\label{sec:appen_setup}
We list hyperparameter settings for the proposed MEST+EM and MEST+EM\&S in Table~\ref{tab:appen_exp_hyperparas}.

\begin{table*}[h]
\centering
\vspace{-1em}
\caption{Hyperparameter settings.}
\scalebox{0.73}{
\begin{tabular}{p{5cm} | p{4cm} | p{4cm} | p{4cm}}
\toprule
Experiments & VGG-19 on CIFAR & ResNet-32 on CIFAR & ResNet-50/34 on ImageNet \\ \hline \hline
\multicolumn{4}{c}{Regular hyperparameter settings} \\ \midrule
Training epochs ($\tau_{end}$) & 160 & 160 & 150 \\ \midrule
Batch size & 64 & 64 & 2048 \\ \midrule
Learning rate scheduler & cosine & cosine & cosine\\ \midrule
Initial learning rate & 0.1 & 0.1 & 2.048 \\ \midrule
Ending learning rate & 4e-8 & 4e-8 & 0 \\ \midrule
Momentum & 0.9 & 0.9 & 0.875 \\ \midrule
$\ell_{2}$ regularization  & 5e-4 & 1e-4 & 3.05e-5 \\ \midrule
Warmup epochs & 5 & 0 & 8 \\ \hline \hline
\multicolumn{4}{c}{MEST hyperparameter settings} \\ \midrule
Number of epochs do EM ($\tau_{stop}$) & 130 & 130 & 120\\ \midrule
EM frequency ($\Delta \tau$) & 5 & 5 & 2 \\ \midrule
MEST+EM schedule & 0 - 100:  \hspace{17pt} \texttt{RM} (s + 0.05)  & 0 - 100:  \hspace{17pt} \texttt{RM} (s + 0.05) & 0 - 90:  \hspace{21pt} \texttt{RM} (s + 0.05) \\
\texttt{ArgRemoveTo} sparsity (\texttt{RM}) &  \hspace{51pt} \texttt{GR} (s)  & \hspace{51pt} \texttt{GR} (s) &\hspace{51pt} \texttt{GR} (s) \\
\texttt{ArgGrowTo} sparsity (\texttt{GR}) & 100 - 130:\hspace{10pt} \texttt{RM} (s + 0.025)  & 100 - 130:\hspace{10pt} \texttt{RM} (s + 0.025) & 90 - 120:\hspace{14pt} \texttt{RM} (s + 0.025) \\
& \hspace{51pt} \texttt{GR} (s)  & \hspace{51pt} \texttt{GR} (s) &\hspace{51pt} \texttt{GR} (s) \\
& 130 - 160: \hspace{8pt} No EM  & 130 - 160: \hspace{8pt} No EM &  120 - 150: \hspace{8pt} No EM\\ \midrule
MEST+EM\&S schedule & 0 - 100: \hspace{16pt} \texttt{GR} (s - 0.05)  & 0 - 100: \hspace{16pt} \texttt{GR} (s - 0.05) & 0 - 90: \hspace{21pt} \texttt{GR} (s - 0.05)\\
\texttt{ArgRemoveTo} sparsity (\texttt{RM})&  \hspace{51pt} \texttt{RM} (s)  & \hspace{51pt} \texttt{RM} (s) & \hspace{51pt} \texttt{RM} (s)\\
\texttt{ArgGrowTo} sparsity (\texttt{GR}) & 100 - 130: \hspace{7pt} \texttt{GR} (s - 0.025)  & 100 - 130: \hspace{7pt} \texttt{GR} (s - 0.025) & 90 - 120: \hspace{11pt} \texttt{GR} (s - 0.025)\\
& \hspace{51pt} \texttt{RM} (s)  & \hspace{51pt} \texttt{RM} (s) & \hspace{51pt} \texttt{RM} (s)\\
& 130 - 160: \hspace{7pt} No EM  & 130 - 160: \hspace{7pt} No EM  &120 - 150: \hspace{7pt} No EM \\ \midrule
Importance coefficient ($\lambda$) & 0.01 & 0.01 & 0.001 \\

\bottomrule
\end{tabular}}
\vspace{-1em}
\label{tab:appen_exp_hyperparas}
\end{table*}

\section{Accuracy Results of ResNet-32 on CIFAR-10, VGG-19 on CIFAR-10 and CIFAR-100, and ResNet-34 on ImageNet-2012}
\label{sec:appen_other_network}
\vspace{-1em}

\begin{figure}[h]
    \centering
    \includegraphics[width=1 \textwidth]{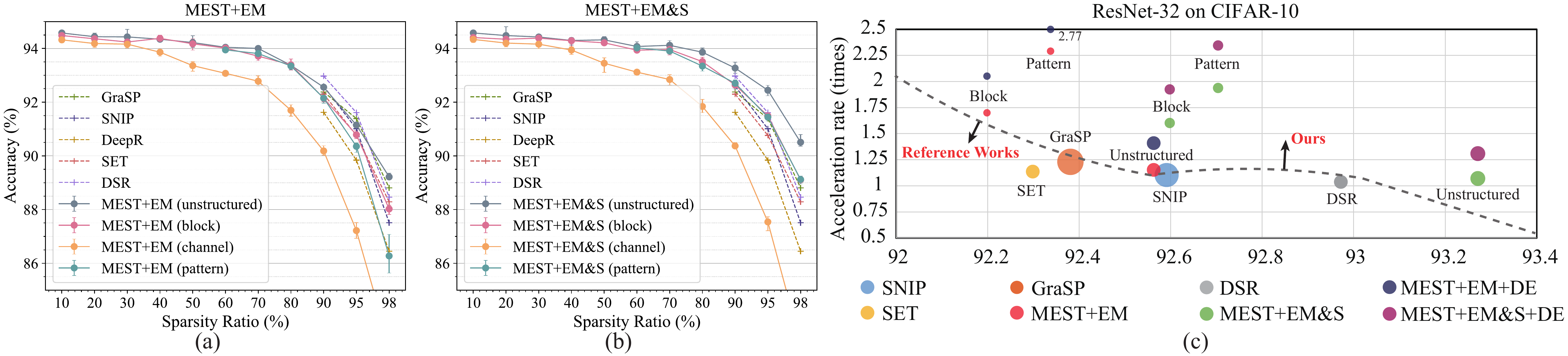}
    \caption{Results obtained using ResNet-32 on CIFAR-10. We choose the reference works that can maintain both sparse weights and gradients along the entire training process. (a) and (b) Accuracy results of the proposed MEST framework using different sparsity schemes and sparsity. (c) Comparison with representative SOTA works in accuracy, training acceleration rate, and memory footprint. Sparsity ratio is 90\% for all results. The acceleration rate is normalized with respect to dense training. The size of the circles represents the relative cost of the memory footprint. We use $th=0$ as the threshold for DE.}
    \label{fig:appen_acc_acceleration_res32_cf10}
\end{figure}

\begin{figure*}[h!]
	\centering
	\begin{minipage}[b]{0.99\textwidth}
		\subfigure[Accuracy comparison on VGG-19 using CIFAR-10 dataset.]{
			\includegraphics[width=0.45\textwidth]{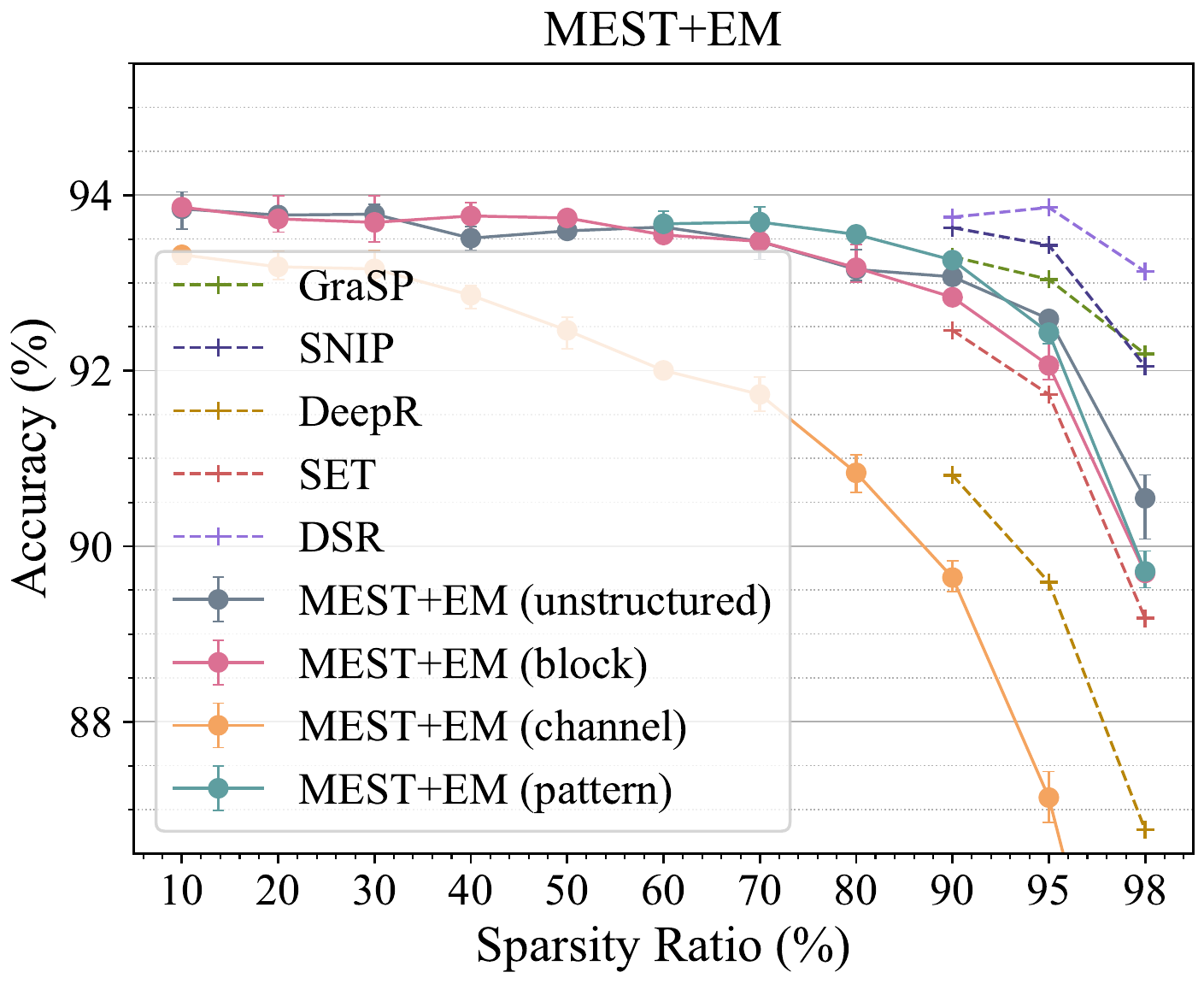} 
			\includegraphics[width=0.45\textwidth]{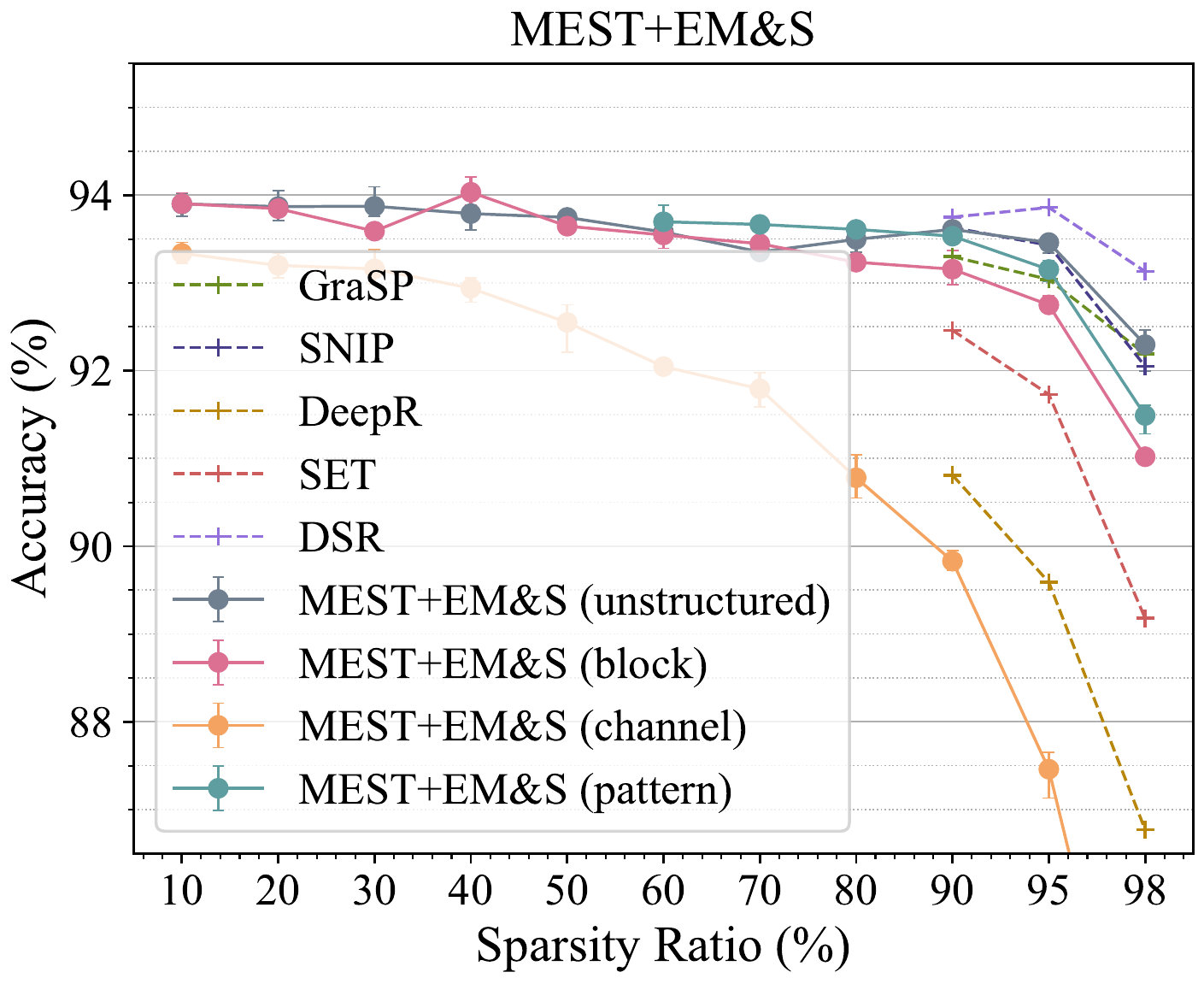}
			\label{fig:vgg19_acc_vs_prune_cifar10}
	}
	\end{minipage}
	
	\begin{minipage}[b]{0.99\textwidth}
		\subfigure[Accuracy comparison on VGG-19 using CIFAR-100 dataset.]{
			\includegraphics[width=0.45\textwidth]{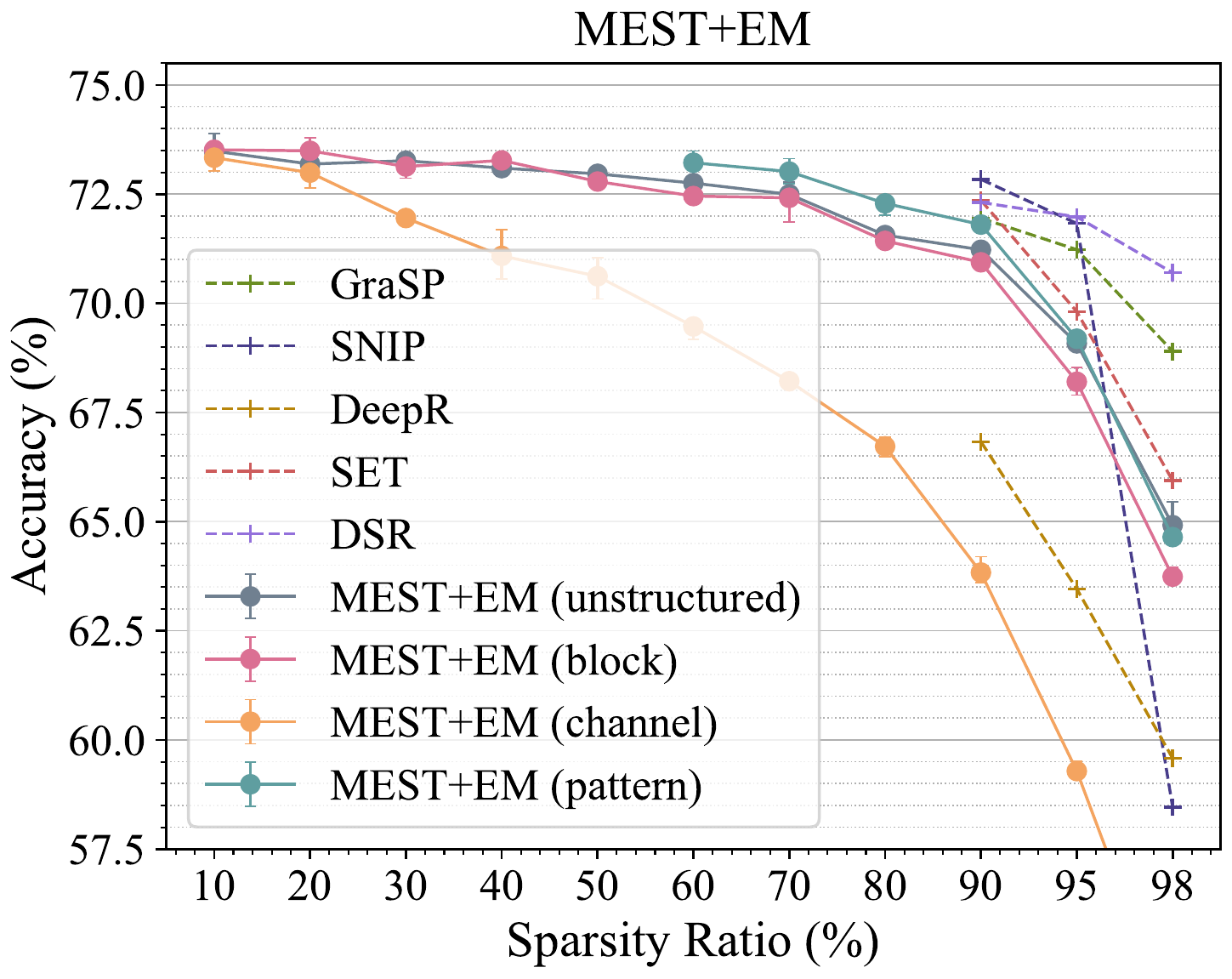} 
			\includegraphics[width=0.45\textwidth]{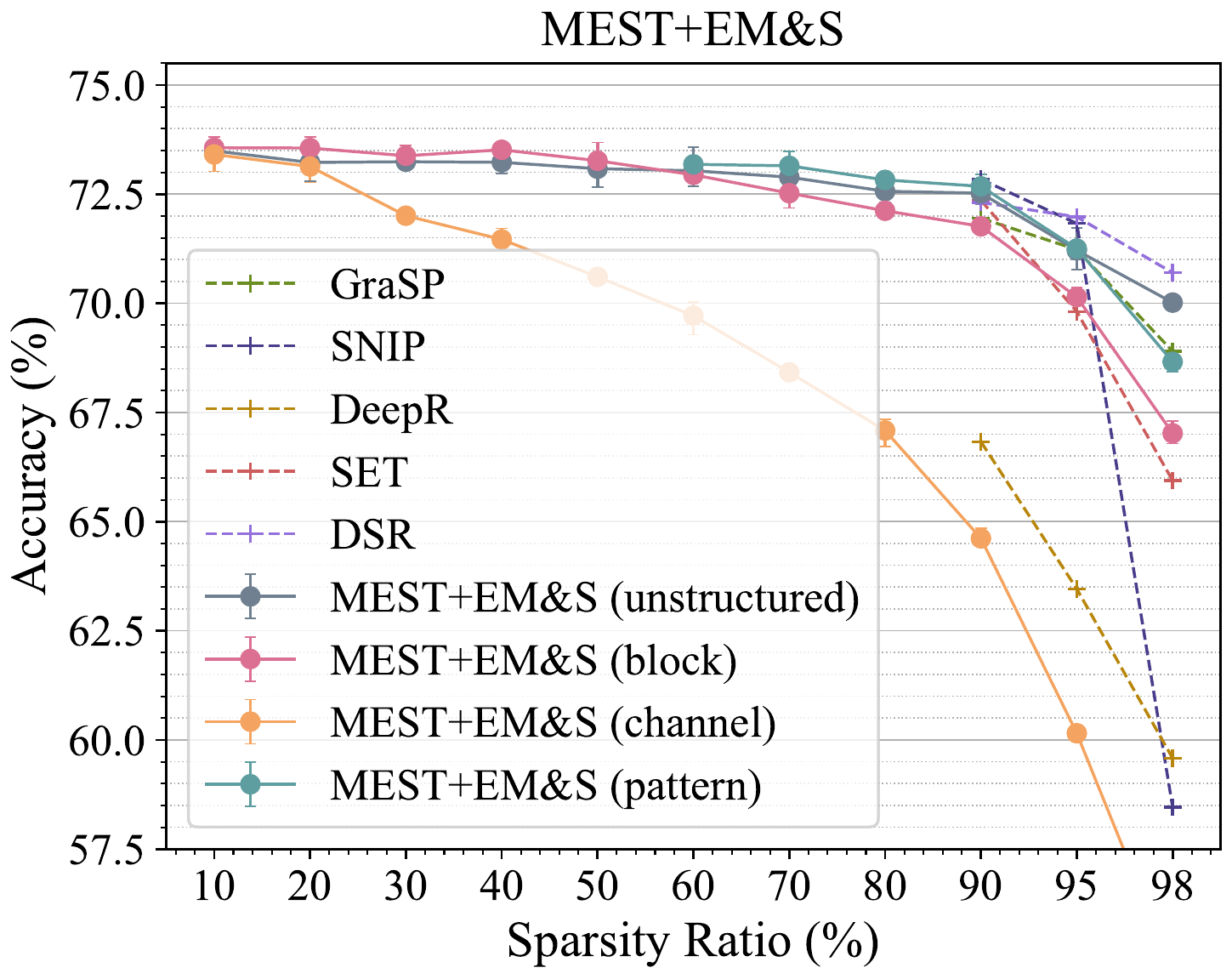}
			\label{fig:vgg19_acc_vs_prune_cifar100}
	}
	\end{minipage}
	\caption{Accuracy of the proposed MEST framework using different sparsity schemes on VGG-19.}
	\label{fig:vgg19_acc_vs_prune_cifar}
\end{figure*}

The MEST accuracy results for ResNet-32 on CIFAR-10 is shown in Figure~\ref{fig:appen_acc_acceleration_res32_cf10}. When incorporating the data-efficient (DE) training, $29.5\%\sim30\%$ training examples are removed without decreasing the accuracy.
We also validate our MEST on VGG-19 using CIFAR-10 and CIFAR-100.
The results are shown in Table~\ref{tab:cifar_vgg19_results} and Figure~\ref{fig:vgg19_acc_vs_prune_cifar}. 
For the ImageNet dataset, we also show MEST accuracy results on ResNet-34.
Since the size of the ImageNet dataset itself is about 150GB and a ResNet50 requires more than 1 day to train on a 8 x 2080Ti GPU server, it is impractical to be trained on current edge devices such as mobile phones. 
Therefore, we mainly use ImageNet dataset to show the accuracy results and validate the effectiveness of our MEST.
Table~\ref{tab:imagenet_resnet34_results} shows the accuracy results with both regular training effort (i.e., 150 epochs) and 1.7$\times$ effort  (i.e., 250 epochs). 
When incorporating the data-efficient (DE) training, $9.3\%\sim10.7\%$ training examples are removed.


\begin{table*}[t]
\centering
\vspace{-1em}
\caption{Accuracy comparison with SOTA works using VGG-19 on CIFAR-10 and CIFAR-100.}
\scalebox{0.70}{
\begin{tabular}{l | c | ccc | ccc}
\toprule
Dataset    & Memory     & \multicolumn{3}{c}{CIFAR-10}     & \multicolumn{3}{|c}{CIFAR-100}    \\ 
  & Footprint      &
\multicolumn{3}{c}{(\emph{Dense: 94.20})}     & \multicolumn{3}{|c}{(\emph{Dense: 74.17})}    \\
\midrule
Sparsity ratio  &        & 90\%  & 95\%  & 98\%  &  90\%  & 95\%  & 98\%  \\ \midrule
LT~\cite{frankle2018lottery}    & dense  & 93.51 & 92.92 & 92.34  & 72.78 & 71.44 & 68.95 \\ \midrule
SNIP~\cite{lee2018snip}   & dense  & 93.63 & 93.43 & 92.05 & 72.84 & 71.83 & 58.46 \\
GraSP~\cite{wang2019picking}     &  dense   & 93.30 & 93.04 & 92.19  & 71.95 & 71.23 & 68.90 \\ \midrule
DeepR~\cite{bellec2018deep}     &   sparse   & 90.81 & 89.59 & 86.77 & 66.83 & 63.46  & 59.58 \\
SET~\cite{mocanu2018scalable}  & sparse & 92.46 & 91.73 & 89.18  & 72.36 & 69.81 & 65.94 \\
DSR~\cite{mostafa2019parameter} & sparse & 93.75 & 93.86 & 93.13 & 72.31 & 71.98  & 70.70  \\ \midrule
MEST (vanilla)  &  sparse  & 91.73$\pm$0.27  & 90.43$\pm$0.33 & 87.82$\pm$0.27 & 68.57$\pm$0.38 & 65.01$\pm$0.32 & 60.88$\pm$0.36 \\
MEST+EM    &  sparse  & 93.07$\pm$0.49 & 92.59$\pm$0.36 & 90.55$\pm$0.37 & 71.23$\pm$0.33 & 69.08$\pm$0.46 & 64.92$\pm$0.42 \\
MEST+EM\&S & sparse & 93.61$\pm$0.36 & 93.46$\pm$0.41 & 92.30$\pm$0.44 & 72.52$\pm$0.37 & 71.21$\pm$0.41 & 69.02$\pm$0.34 \\ \bottomrule
\end{tabular}}
\label{tab:cifar_vgg19_results}
\end{table*}

\begin{table}[h]
\centering
\caption{Accuracy with ResNet-34 on ImageNet.}
\scalebox{0.88}{
    \begin{tabular}{l | cc  cc}
        \toprule
        Method &  \multicolumn{4}{c}{Top-1 accuracy (\%)}   \\ \midrule
        Dense & \multicolumn{4}{c}{74.08}   \\ \midrule
        Sparsity ratio & \multicolumn{1}{c}{60\%} &\multicolumn{1}{c}{70\%} &\multicolumn{1}{c}{80\%} & \multicolumn{1}{c}{90\%} \\ \midrule
        MEST (vanilla) &  73.08 & 70.71 & 69.74 &  65.68   \\
        MEST+EM & 74.10 & 73.66 & 72.83 & 70.38   \\
        MEST+EM\&S &  74.12 & 73.81 & 73.57 & 72.10   \\
        MEST+EM\&S+DE &  74.17 & 73.83 & 73.48 & 72.03  \\
        \hline \hline
        MEST$_{1.7\times}$ (vanilla) & 73.21  & 70.79 & 69.92  &  65.77   \\
        MEST$_{1.7\times}$+EM & 74.30  & 73.89 &  73.12 &  70.76  \\
        MEST$_{1.7\times}$+EM\&S & 74.34  & 73.97 & 73.86 &  72.25   \\
        MEST$_{1.7\times}$+EM\&S+DE & 74.37  & 73.93 & 73.79 &  72.13   \\ 
        \bottomrule
    \end{tabular}
}
\label{tab:imagenet_resnet34_results}
\end{table}

\newpage

\section{MEST Accuracy on Compact Model MobileNet-V2 and Deeper Model ResNet-110}
\label{sec:compare_mbnet_res110}

We also evaluate our MEST+EM\&S on MobileNet-V2 as the representative of compact models and on ResNet-110 as the representative of deep models.
Table~\ref{tab:comp_acc_mbv2_res110} shows the accuracy achieved by our MEST+EM\&S method on CIFAR10 under different sparsity ratios, and Table~\ref{tab:comp_flops_mbv2_res110} shows the overall training FLOPs ($\times$e12) and number of model parameters (M). 
The MobileNet-V2 is more sensitive to sparsity compared to ResNet-32 and ResNet-110, as shown in Table~\ref{tab:comp_acc_mbv2_res110}. 
Actually, the ResNet-32 (as well as ResNet-20, ResNet-110) are the lightweight ResNet version dedicated to CIFAR tasks, while the ResNet-18, ResNet-34, and ResNet-101 are the large versions for the ImageNet. 
So, as we can see from Table~\ref{tab:comp_flops_mbv2_res110}, the ResNet-32 is even smaller than MobileNet-V2 (1.86M parameters v.s. 2.3M parameters.), while the computation cost of ResNet-32 is higher than MobileNet-V2 due to the depth-wise separable CONV. Note that the ResNet-32 here (and in the paper) is a (2$\times$) widened version, which is consistent with the reference works cited in the paper (all as shown in Table~\ref{tab:cifar_results}). This is the reason that the number of parameters and training cost of ResNet-32 is similar to the ResNet-110 as shown in Table~\ref{tab:comp_flops_mbv2_res110}.

It is interesting to see that, under 90\% sparsity, ResNet-32 has a similar accuracy and training FLOPs as the MobileNet-V2 under 60\% sparsity, while the number of parameters of ResNet-32 is 4.8$\times$ less than MobileNet-V2. 
For this case, the ResNet-32 will be more desired than MobileNet-V2. Moreover, MobileNet-V2 is much deeper (57 CONV layers) than ResNet-32, which will require more data movement among memory and cache for reading and writing intermediate results and lead to a higher execution overhead.

\begin{table*}[h]
\centering
\caption{Accuracy comparison on ResNet-32, MobileNet-V2, and ResNet-110 on CIFAR-10 using MEST+EM\&S.}
\begin{tabular}{|l|c|ccccc|}
\hline
Sparsity & Dense & 50\% & 60\% & 70\% & 80\% & 90\% \\ \hline
ResNet-32 & 94.88 & 94.41 & 94.05 & 94.14 & 93.70 & \textbf{93.27} \\ \hline
MobileNet-V2 & 94.08 & 94.06 & \textbf{93.32} & 93.05 & 92.38 & 90.61 \\ \hline
ResNet-110 & 94.64 & 93.47 & 93.73 & 93.62 & 93.26 & \textbf{92.29} \\ \hline
\end{tabular}
\label{tab:comp_acc_mbv2_res110}
\end{table*}

\begin{table*}[h!]
\centering
\caption{Comparison of train FLOPs ($\times$e12) and number of parameters (M) on ResNet-32, MobileNet-V2, and ResNet-110 on CIFAR-10 using MEST+EM\&S.}
\scalebox{0.95}{
\begin{tabular}{|l|c|ccccc|}
\hline
Sparsity & Dense & 50\% & 60\% & 70\% & 80\% & 90\% \\ \hline
ResNet-32 & 6.38 / 1.86 & 3.30 / 0.93 & 2.68 / 0.74 & 2.07 / 0.56 & 1.45 / 0.37 & \textbf{0.83 / 0.19} \\ \hline
MobileNet-V2 & 2.11 / 2.30 & 1.09 / 1.15 & \textbf{0.88 / 0.92} & 0.68 / 0.69 & 0.48 / 0.46 & 0.28 / 0.23 \\ \hline
ResNet-110 & 5.74 / 1.70 & 2.97 / 0.85 & 2.41 / 0.68 & 1.86 / 0.51 & 1.31 / 0.34 & \textbf{0.75 / 0.17} \\ \hline
\end{tabular}
}
\label{tab:comp_flops_mbv2_res110}
\end{table*}

\section{Why Does Memory-Economic Critical for Training on Edge Devices?}
\label{sec:appen_why_mem_eco_critical}

The availability of edge devices for training requires consideration of two aspects:
(1) whether the dataset and model can be accommodated by a mobile device; 
(2) whether the free space of device memory (RAM) is sufficient for the required training memory footprint. 

The current mobile devices generally have memory in GB levels. For example, current general mobile devices such as Samsung Galaxy A20s, Google Pixel 3, and Samsung S20 have 2GB or more memory.
However, unlike the training on a high-end GPU cluster where all the memory can be reserved for training, the memory on mobile devices will also be partially occupied by the operating system and other backend applications. 
This puts an even greater strain on the memory of mobile devices. 
Thus, having a smaller memory footprint will always benefit the training on mobile devices. 

We use ResNet-32 and VGG-19 as examples. Assume 32 bits weights, 8 bits indices are used with batch size of 64.
For ResNet-32, the dense model and the methods that involve dense computations require a 462MB memory footprint for model size, while our sparse model with unstructured sparsity requires 46MB (69MB) and 23MB (46MB) under 90\% and 95\% sparsity for MEST+EM(\&S), respectively.
For VGG-19, a significant reduction can be obtained, where the dense model requires 4964MB memory footprint, while our sparse model with unstructured sparsity requires 494MB (744MB) and 247MB (494MB) under 90\% and 95\% sparsity for MEST+EM(\&S), respectively.

MEST successfully reduces the memory footprint (to less than 100M and 800M for ResNet-32 and VGG-19) while maintaining a similar or higher accuracy than prior works, while the reference methods either require a dense memory footprint (LT, SNIP, GraSP, RigL) or suffer a severer accuracy degradation.

\section{Layer-wise Sparsity Scheme and Sparsity Ratio}\label{sec:appen_layerwise_scheme_and_ratio}

As shown in the main paper, different sparsity schemes have different performance in accuracy and acceleration rate.
Moreover, pattern-based sparsity is only applicable to 3$\times$3 CONV layers, while many popular networks contain a large portion of 1$\times$1 CONV layers or FC layers.
Therefore, hybrid sparsity schemes may be a better option, although the pattern-based sparsity is still preferred for 3$\times$3 CONV layers.
On the other hand, different types and different sizes of layers inherently exhibit different weight redundancy and therefore deserve  non-uniform sparsity ratios among layers.

In this section, we further investigate the performance when using a layer-wise sparsity scheme and sparsity ratio assignment.
We use ResNet-50 on CIFAR-100 as an example, and Table~\ref{tab:appen_hybrid} shows the results of accuracy and training speed when using a single sparsity scheme or using different sparsity schemes on different types of layers. The training speed is the time of a training iteration in seconds while using a batch size of 64 and measured on a Samsung smartphone using the mobile GPU.
For ResNet-50, over 50\% of weights and computations are contributed by the 1$\times$1 CONV layers. If we only adopt pattern-based sparsity, which is only applicable to 3$\times$3 CONV layers, the overall sparsity ratio is 44\% under a 90\% sparsity on the 3$\times$3 CONV layers.
Therefore, only use pattern-based sparsity is not able to achieve a high sparsity ratio and hence lower training acceleration.
Block-based sparsity can be adopted across the entire network. But both the accuracy and training speed are lower than using hybrid sparsity schemes, i.e., adopting pattern-based sparsity to 3$\times$3 CONV layers and block-based sparsity to 1$\times$1 CONV layers, respectively.

We also explore different sparsity ratio strategies by comparing three sparsity ratio settings, including 1) the uniform sparsity ratio (90\%) on all the layers, 2) a fixed ratio (1.12:1) between 3$\times$3 CONV layers and 1$\times$1 CONV layers in the entire network (i.e., 95\% pattern-based sparsity for
all 3$\times$3 CONV layers and 85\% block-based sparsity for 1$\times$1 CONV layers), and 3) a layer-wise ratio assignment proportional to the layer size. 
All three strategies are hybrid sparsity schemes and have the same overall sparsity ratio (90\%).

\begin{table}[h!]
\centering
\caption{Comparison of accuracy and training speed using different sparsity schemes and different layer-wise sparsity ratios on ResNet-50 using CIFAR-100.}
\scalebox{0.84}{
\begin{tabular}{l  c  c  c}
\toprule

\multirow{2}{*}{Scheme} & \multicolumn{1}{|c}{Sparsity} & \multicolumn{1}{|c}{Accuracy} & \multicolumn{1}{|c}{Training}   \\ 
& \multicolumn{1}{|c}{Ratio} & \multicolumn{1}{|c}{(\%)} & \multicolumn{1}{|c}{Speed ($s/iter$)} \\ \midrule
Dense & 0\% &77.18  & 11.92\\ \midrule
\multicolumn{4}{c}{MEST+EM} \\ \midrule
Pattern & 44\% (90\%) &75.36 & 8.18\\
Block & 90\% &72.82 & 6.25\\
Hybrid (uniform) & 90\% &72.87 & 5.39\\
Hybrid (1.12:1) & 90\% &73.24 & 5.55\\
Hybrid (proportional) & 90\% &73.56 & 5.62\\ \midrule 
\multicolumn{4}{c}{MEST+EM\&S} \\ \midrule
Pattern & 44\% (90\%) &75.88 & 8.36\\
Block & 90\% &73.68 & 6.79\\
Hybrid (uniform) & 90\% &73.72 & 5.99\\
Hybrid (1.12:1) & 90\% &73.98& 6.08\\
Hybrid (proportional) & 90\% &74.12 & 6.15\\

\bottomrule
\end{tabular}}
\label{tab:appen_hybrid}
\end{table}

It can be observed that the three strategies have similar training speed, where the uniform sparsity ratio is the fastest since the acceleration rate is not linearly increased along with the increased sparsity ratio, as shown in Figure \ref{fig:speedup} in the main paper.
On the other hand, when using non-uniform sparsity ratios, the accuracy can be improved, which indicates the larger layers have more redundant weights compared to smaller layers, and can tolerant a higher sparsity ratio.

\section{Combinations of Dataset Compression and Model Sparsity}
\label{sec:appen_ablation_combination}
In our work, we intend to incorporate dataset-efficient training on top of the sparse training and without further decreasing the accuracy.
However, when a minor accuracy drop is allowed, selecting the best-suited combination of dataset compression and model sparsity to achieve a higher acceleration while maintaining a higher accuracy is an interesting topic that can be further studied.

\begin{table}[h]
\centering
\caption{Comparison of accuracy results on different dataset compression and model sparsity combinations. The results are obtained by using MEST+EM\&S with unstructured sparsity on ResNet-32 and CIFAR-10 dataset.}
\begin{tabular}{|l|c|c|c|}
\hline
Scheme & baseline & \ding{172} & \ding{173} \\ \hline
Sparsity & 90\% & 95\% & 90\% \\ \hline
Removed examples & 0 & 0 & 17900 \\ \hline
Phase-1 epochs & - & - & 40 \\ \hline
Final accuracy (\%) & 93.27 & 92.44 & 93.02 \\ \hline
\end{tabular}
\label{tab:appen_combine}
\end{table}

For example, as shown in Table~\ref{tab:appen_combine}, we consider the MEST+EM\&S result under 90\% unstructured sparsity as the baseline scheme. If we intend to further increase the acceleration rate, we can choose two different schemes, including \ding{172} further increasing the sparsity or \ding{173} incorporating data-efficient training (i.e., compress the dataset).
Based on our measurements, the scheme \ding{172} and \ding{173} provide the same acceleration rate (1.27$\times$) on top of the baseline scheme. 
But we observe that incorporating data-efficient training instead of further increasing the model sparsity can deliver higher accuracy.
Therefore, we may hypothesize that when the model sparsity goes beyond a certain degree, to further increase the acceleration rate while preserving a higher accuracy, it is more desired to compress the dataset than compress the model.
Since both the number of removed examples and the removing epoch will affect the final model accuracy, it is a complicated problem that is worth to be further studied in the future.

\end{document}